\newif\ifconfver
\confvertrue        

\documentclass[10pt,twocolumn,twoside]{IEEEtran}

\ifCLASSOPTIONcompsoc
  \usepackage[nocompress]{cite}
\else
  \usepackage{cite}
\fi
\usepackage{hyperref}

\usepackage{calc,amsfonts,amssymb,amsmath,bm,url,color,theorem,graphicx,cite}
\usepackage{psfrag,subfigure,float}
\usepackage{multirow,subfigure,amsmath,epsfig,amsfonts,amssymb,psfig,graphics,psfrag,theorem,calc,subfigure,url,bm,cite}
\usepackage{stfloats,hyperref}  
\usepackage{algorithm,algorithmic}
\usepackage{diagbox} 
\usepackage{hhline}
\usepackage{caption}
\usepackage{booktabs}
\usepackage{mathtools}

\usepackage[table,xcdraw]{xcolor}
\usepackage{colortbl} 

\makeatletter
\def\multilimits@{\bgroup
	\Let@
	\restore@math@cr
	\default@tag
	\baselineskip\fontdimen10 \scriptfont\tw@
	\advance\baselineskip\fontdimen12 \scriptfont\tw@
	\lineskip\thr@@\fontdimen8 \scriptfont\thr@@
	\lineskiplimit\lineskip
	\vbox\bgroup\ialign\bgroup\hfil$\m@th\scriptstyle{##}$\hfil\crcr}
\def\Sb{_\multilimits@}
\def\endSb{\crcr\egroup\egroup\egroup}
\makeatother

\definecolor{orange}{RGB}{255,107,0}

\def\red{\color{red}}

{\begin{list}{}%
		{\setlength{\rightmargin}{0pc}%
			\setlength{\leftmargin}{1pc} }} %
	{\end{list}}

\newlength{\twidth}
\ifconfver
\else
\fi

\definecolor{mediumseagreen}{rgb}{0.58, 0.44, 0.86}
\definecolor{orange}{RGB}{255,107,0}

\def\red{\color{red}}

\def\red{\color{red}}

\theorembodyfont{\rmfamily}

{\begin{list}{}{
			\settowidth{\labelwidth}{\mbox{\textnormal{#1}}}%
			\setlength{\leftmargin}{\labelwidth+\labelsep}}}%
	{\end{list}}

\newcommand\bW{\ensuremath{{\bm W}}}

\newcommand\bR{\ensuremath{{\bm R}}}

\newcommand\bX{\ensuremath{{\bm X}}}
\newcommand\bZ{\ensuremath{{\bm Z}}}

\newcommand\bU{\ensuremath{{\bm U}}}

\ifconfver

\author{Chih-Chung Hsu,~\IEEEmembership{Senior Member,~IEEE}, Chih-Yu Jian, Eng-Shen Tu, Chia-Ming Lee and Guan-Lin Chen \ \red{\tt\url{https://rcts01.github.io/}}}

\title{Real-Time Compressed Sensing for Joint Hyperspectral Image Transmission and Restoration for CubeSat

	\thanks{This study was supported in part by the National Science and Technology Council (NSTC), Taiwan, under grants MOST 112-2221-E-006 -157 -MY3 and 111-2221-E-006 -210; partly by the Higher Education Sprout Project of Ministry of Education (MOE) to the Headquarters of University Advancement at National Cheng Kung University (NCKU). We thank to National Center for High-performance Computing (NCHC) of National Applied Research Laboratories (NARLabs) in Taiwan for providing computational and storage resources.}
\thanks{\textit{(Corresponding author: Chih-Chung Hsu.)}}
\thanks{
		C.-C. Hsu, C.-Y. Jian, C.-M. Lee and G.-L. Chen are with Institute of Data Science and Department of Statistics, Miin Wu School of Computing, National Cheng Kung University, Tainan, Taiwan (R.O.C.), (e-mail:cchsu@gs.ncku.edu.tw, zuw408421476@gmail.com, alright1117@gmail.com)
		E.-S. Tu is with Department of Computer Science, National Cheng Kung University, Tainan, Taiwan (R.O.C.)}
}

\else

\fi

\begin{document}
	
	\maketitle
 
	\ifconfver \else \vspace{-0.5cm}\fi

        \begin{abstract}

            This paper addresses the challenges associated with hyperspectral image (HSI) reconstruction from miniaturized satellites, which often suffer from stripe effects and are computationally resource-limited. We propose a Real-Time Compressed Sensing (RTCS) network designed to be lightweight and require only relatively few training samples for efficient and robust HSI reconstruction in the presence of the stripe effect and under noisy transmission conditions. 
            The RTCS network features a simplified architecture that reduces the required training samples and allows for easy implementation on integer-8-based encoders, facilitating rapid compressed sensing for stripe-like HSI, which exactly matches the moderate design of miniaturized satellites on push broom scanning mechanism. This contrasts optimization-based models that demand high-precision floating-point operations, making them difficult to deploy on edge devices. Our encoder employs an integer-8-compatible linear projection for stripe-like HSI data transmission, ensuring real-time compressed sensing. Furthermore, based on the novel two-streamed architecture, an efficient HSI restoration decoder is proposed for the receiver side, allowing for edge-device reconstruction without needing a sophisticated central server. This is particularly crucial as an increasing number of miniaturized satellites necessitates significant computing resources on the ground station. 
            Extensive experiments validate the superior performance of our approach, offering new and vital capabilities for existing miniaturized satellite systems. 
            
		\bfseries{\em Index Terms---}
		deep learning, 
		hyperspectral image, 
		compressed sensing,
		hyperspectral restoration,
		real-time applications.
	\end{abstract}
	
	\ifconfver \else \vspace{-0.0cm}\fi

	\ifconfver \else \vspace{-0.5cm}\fi

	\ifconfver \else  \fi

\section{Introduction}
    
Hyperspectral remote sensing benefits tasks including environmental monitoring, matter recognition, urban planning, agriculture, and surveillance. Deep learning (DL), growing in popularity for vision tasks, has significantly impacted hyperspectral remote sensing applications \cite{dcsn, fusion1, fusion2, fusion3}, demonstrating superior performance over conventional optimization-based approaches \cite{keshava2002spectral,BPCSNC2013,Stein2002,chang2010real,lin2018maximum}.
Hyperspectral images (HSI) contain hundreds of spectral bands, both visible and invisible, resulting in a substantial data volume. Efficient remote sensing is essential for transmitting HSI to ground receivers for applications in matter recognition, urban planning, and agriculture. The increase in Low Earth Orbit satellites (LEOS) based on CubeSat/miniaturized technology has led to bandwidth challenges due to substantial data transmission. Hyperspectral compressed sensing (HCS) is thus widely used for concurrent sensing and compression of HSI \cite{martin2016hyperspectral, duarte2011kronecker,wang2017compressive, zhang2016locally}.

Hyperspectral sensors capture data in stripe-like sequences, forming HSI in the along-track direction, a method known as pushbroom sensing. However, satellite movement, orography, or IFOV (Instantaneous Field of View) issues mean that spatial correlation between contiguous slices may not always be guaranteed. Thus, we must perform geometric correction and geographic data projection to achieve spatially correlated images. Given its high computational load, this pre-processing is often deferred to post-transmission at the ground station. Moreover, limited memory in miniaturized satellites restricts the stacking of multiple spectral slices. Stripe-like encoding, as proposed in DCSN \cite{dcsn}, offers a solution by encoding fewer slices, emphasizing the need for efficient and precise stripe-like hyperspectral data transmission from miniaturized satellites to ground receivers.

Conventional data compression strategies like JPEG, JPEG2000, and video codecs could theoretically reduce HSI data volume. Several approaches have been proposed, utilizing principal component analysis (PCA) and codecs such as JPEG2000 \cite{JPEG2000_PCA} and X264 \cite{X264_PCA}. However, given the prevalent use of pushbroom sensing for acquiring HSI (stripe-like data) and the limited storage of miniaturized satellites, traditional image/video compression techniques may lack the necessary computational resources \cite{JPEG2000_PCA, X264_PCA}, rendering these methods incompatible for real-time compressed sensing in such satellites.

The power and computational limitations in CubeSats necessitate the adoption of lightweight and efficient HCS. HCS can perform concurrent compression and sensing, significantly easing computational and storage demands in miniaturized satellites. Therefore, an ultra-lightweight HCS is crucial. Spectral compressive acquisition (SpeCA) \cite{martin2016hyperspectral} leverages spatial and spectral redundancies to create random projection matrices for low-dimensional data extraction. Novel HCS methods utilizing $L_1$-norm and total-variation (TV) regularization have been introduced \cite{duarte2011kronecker,wang2017compressive} to exploit global spatial-spectral correlations in HSI data while maintaining local smoothness. A spatial/spectral compressed encoder (SPACE), based on low-rank representation \cite{space}, transforms HCS into an image fusion process by separately compressing spatial and spectral data. The recently improved SPACE variant, All-Addition Hyperspectral Compressed Sensing (AAHCS), enhances both spectral and spatial qualities of reconstructed HSI data \cite{aahcsd}. However, these optimization-based HCS methods often require lengthy computational times due to the necessity of eigendecomposition. Although AAHCS demonstrates fast sensing and high performance through all-addition operations in the encoder \cite{aahcsd}, real-time compressed sensing is hampered by the required eigendecomposition, which is computationally intensive ($\mathcal{O}(n^3)$) and thus challenging for on-board processing in miniaturized satellites. Moreover, the stripe-like HSI acquired through pushbroom sensing in moderate satellites does not align well with these optimization-based HCS designs, potentially suppressing performance.

Deep learning offers an effective and efficient solution for hyperspectral compressed sensing (HCS) challenges. The recently introduced Deep Compressed Sensing Network (DCSN) addresses stripe-like HSI data processing challenges in miniaturized satellites \cite{dcsn}. DCSN features a non-square kernel within a three-layer convolutional neural network (CNN), creating a lightweight encoder that swiftly compresses HSI data before transmission to the ground station. For data recovery, DCSN employs a complex decoder leveraging Multi-scale Feature Fusion Block (MFB) and Aggregation (MFA).
Despite DCSN's state-of-the-art performance at lower sampling rates (approximately 1\%), achieving real-time HCS is hindered by delays from multi-layer CNNs with non-linear activation functions. Conventional activation functions, such as ReLU, demand additional power and computational resources \cite{act1}. The hardware implementation of these functions requires multiple units to approximate non-linear models, leading to increased computation time \cite{act1,act2,act3}. Consequently, the LeakyReLU \cite{leakyrelu} activation function used in DCSN's encoder may not be practical for resource-constrained devices, like edge devices \cite{ascia2019analyzing,wang2019shenjing}.
 
Minimizing the computational complexity of HSI data decoding is crucial. The surge in edge computing, which involves processing on mobile or low-power embedded devices, is increasingly popular due to its ability to reduce central server load \cite{eg1,eg2,eg3}. Quick reconstruction of compressed hyperspectral data on these devices, including geometric correction and stitching, alleviates decompression burdens on central servers. However, deploying sophisticated decoders like those in DCSN \cite{dcsn} on edge devices is challenging. Additionally, the iterative decoding approach of AAHCS \cite{aahcsd} precludes real-time processing on such devices. As the number of LEOS is expected to rise, simplifying the decoder's computational demands becomes essential for real-time applications.

A critical yet often overlooked issue in hyperspectral compressed sensing (HCS) is the striping effect caused by anomalies in hyperspectral sensor's electrical units \cite{admmadam}. While robust HCS methods like DCSN \cite{dcsn}, AAHCS \cite{aahcsd}, SPACE \cite{space}, and SpeCA \cite{martin2016hyperspectral} typically focus on channel noise during transmission, they seldom address the striping effect. This phenomenon, often handled as a data imputation challenge and addressed through HSI inpainting \cite{admmadam, inpainting1,inpainting2,inpainting3,inpainting4}, is especially prevalent in devices like NASA's EO-1 satellite's Hyperion imager \cite{hyperion}. Addressing this striping effect in HCS is essential for enhancing the robustness of sensing techniques in miniaturized satellites, aiming to minimize stripe artifacts in the reconstructed HSI.

\begin{figure}
    \begin{center}    \includegraphics[width=0.5\textwidth] {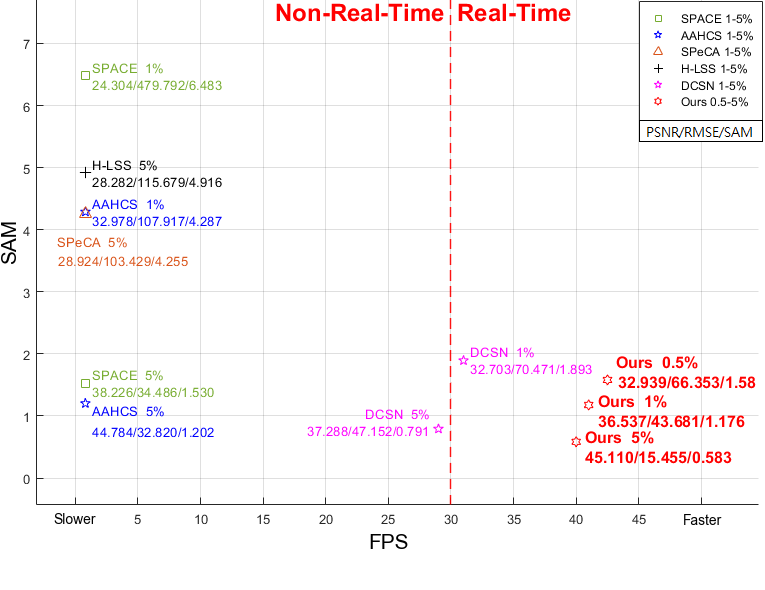}
    \end{center}
    \caption{The performance and computational complexity (decoding and encoding) comparison between the proposed RTCS Network and other peer methods in terms of frame-per-second (FPS) and Spectral angle mapper (SAM), Peak-Signal-to-Noise (PSNR), and Root-Mean-Squared-Error (RMSE), where the DCSN \cite{dcsn} and our RTCS are evaluated in Jetson TX2, the other methods are evaluated on a personal computer, and the sampling rate is ranged in from 0.5\% to 5\%.}
    \label{fig:res_header}
\end{figure}

This manuscript presents the Real-time Compressed Sensing (RTCS) network, a system innovatively designed for the efficient transmission and restoration of hyperspectral data. It addresses challenges such as rapid sensing, stripe effect mitigation, and noise-resistant transmission. Figure \ref{fig:res_header} demonstrates how RTCS adeptly balances efficiency and effectiveness in real-time applications. In our evaluations, conducted on the Jetson TX2 platform, RTCS and DCSN \cite{dcsn} displayed notable performance, while other HCS methods were tested on a personal computer due to their incompatibility with Arm-based architectures (refer to Section \ref{sec:experiments}). In contrast to DCSN's \cite{dcsn} reliance on multi-layer nonlinear activations, which can impede real-time processing, RTCS employs a straightforward linear projection. This approach effectively captures essential spatial and spectral features, resonating with the success of linear models in optimization-based HCS \cite{space, aahcsd}. Furthermore, the encoder in RTCS is tailored for hardware efficiency and compatible with quantization, unlike the high-precision requirements of traditional optimization-based methods.

With the burgeoning number of LEOS, real-time decoding on edge devices is becoming increasingly crucial. Deeper networks typically result in slower inference due to inter-layer dependencies. To address this, our proposed decoder features a reduced depth to facilitate faster inference. It incorporates a novel two-streamed network with a Cross-Scale Feature (CSF) module, which efficiently captures a range of features across different scales. We employ grouped convolution to significantly cut down on the number of parameters and computational complexity, taking advantage of the redundancy in HSI data spectra. This innovation enables the decoder to support real-time HSI reconstruction on edge devices. Additionally, we introduce a Spectral Angle Mapper-aware (SAM) Loss, aiming to enhance spectral quality by guiding spectral signature reconstruction. The RTCS, with its task-specific training approach, demonstrates exceptional resistance to stripe effects commonly encountered in miniaturized satellites. Moreover, the unique two-streamed network design, combined with RTCS's streamlined architecture and reduced parameter count, significantly boosting its adaptability and training efficiency for various hyperspectral sensors.

The primary contributions of this paper are as follows:
\begin{itemize}

\item \textbf{Simple Linear Projection and Edge Device Compatibility in encoder}:
RTCS introduces a unique linear projection in its encoder, precisely aligning with the principles of compressed sensing. This projection not only adheres to the integer-8 standard, making it highly suitable for edge devices, but also offers significant power savings and computational reduction, deviating from traditional optimization approaches.

\item \textbf{Innovative Two-Streamed Network in decoder:}
The novel two-streamed network architecture in RTCS employs a lightweight design that effectively balances performance and robustness. This innovative approach enhances the system's capabilities without imposing additional computational complexity, exemplifying an advanced integration of efficiency and functionality.

\item \textbf{Task-Specific Training for Enhanced Capabilities:}
The RTCS leverages a task-specific training policy, enabling it to excel not only in compressed sensing for efficient transmission but also in reducing stripe effects in decoded hyperspectral images. This marks the first instance of seamlessly integrating compressed sensing and tensor inpainting within a single model, achieving unparalleled performance across extensive datasets.

\end{itemize}

The remainder of this paper is organized as follows.
Section \ref{sec:proposed_method} introduces the proposed RTCS architecture design.
Section \ref{sec:experiments} experimentally demonstrates the superiority of RTCS over benchmark methods. Finally, Section \ref{sec:conclusion} presents the conclusions drawn from this study.

	
\section{Proposed Method} \label{sec:proposed_method}
	\begin{figure*}
	\begin{center}
		\includegraphics[width=1\textwidth] 
  {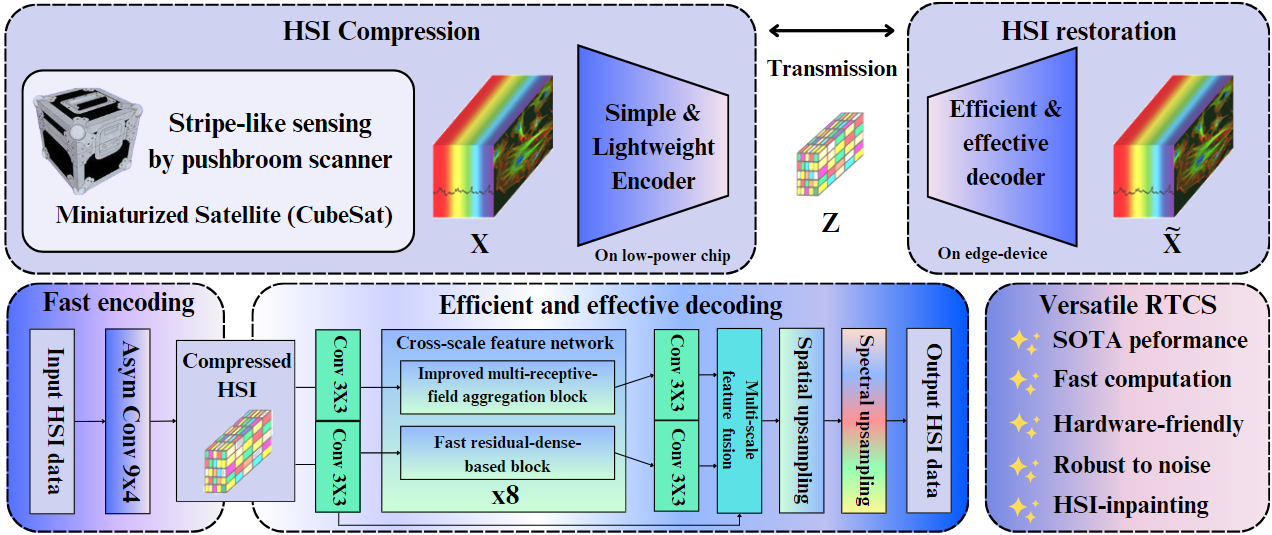}
	\end{center}
	\caption{Overview of the proposed RTCS framework for hyperspectral data processing in CubeSat platforms, illustrating the compression and restoration pipeline. The process includes stripe-like sensing, encoding on a low-power chip, and efficient decoding on edge devices. The system is designed for real-time operation, emphasizing minimizing memory usage and computational load for resource-constrained satellite systems.}
	\label{fig:overview}
    \end{figure*}


As illustrated in Figure \ref{fig:overview}, we present a novel transmission framework designed to minimize the computational complexity inherent in the encoding and decoding processes. This enables the swift decoding of hyperspectral data on edge devices, thereby alleviating the computational burden on central servers. Such a framework gains prominence as the constellation of Low Earth Orbit (LEO) satellites, including CubeSats, expands. It {{allows}} central servers to allocate their computational power more efficiently to demanding {\red{tasks}}, like geometric correction or hyperspectral data integration, rather than spending it on decoding. Thus, the ability to perform encoding and decoding in real-time is crucial.

The Real-Time Compressed Sensing (RTCS) method presented in this study is specifically designed for use in compact satellite systems. It introduces a stripe-based compressed sensing technique that is compatible with the sensing capabilities of modern miniaturized satellites and is optimized for low memory usage. This approach leverages simple matrix operations to facilitate real-time onboard satellite computations. The RTCS's encoder operates through a linear transform-based method that is both hardware-friendly and conducive to integer operations, contrasting with traditional Hyperspectral Compressed Sensing (HCS) that relies on high-precision floating-point operations. As a result, our encoder is well-suited for edge devices with limited processing power. The RTCS's support for stripe-like HCS significantly reduces memory demands on miniaturized satellites, requiring only four hyperspectral stripes for effective compression and immediate transmission. Even on satellites with stringent resource limitations, RTCS maintains its capability to encode hyperspectral data efficiently.


Ground station decoding speed is a critical factor. With an increasing fleet of Low Earth Orbit Satellites (LEOS), the central server may be inundated with compressed hyperspectral data awaiting decoding and integration. The RTCS method we propose circumvents the need for server-centric decoding, instead facilitating the reconstruction of hyperspectral data directly on edge devices within ground stations. This expedites data reconstruction and conserves computational resources for more critical HSI tasks. For this purpose, we introduce a Cross-Scale Feature (CSF) network decoder that employs a dual-stream, shallow architecture, effectively extracting multi-scale features while maintaining computational efficiency. This innovation significantly reduces the workload on central servers.

Moreover, we propose a Spectral Angle Mapper-aware (SAM) Loss to guide the network in preserving spectral signatures, thus enhancing spectral quality. A task-specific training policy is subsequently introduced to improve the robustness of our RTCS, allowing for real-time compressed sensing and reduced stripe effects during the decoding phase. This feature renders the proposed RTCS more practical. Due to the relatively few parameters of the proposed RTCS, relatively few training samples are required to match the promising performance, enhancing feasibility when re-training the RTCS for further applications.

In the ensuing sections, we thoroughly discuss the effectiveness and efficiency of the proposed encoder and decoder in our RTCS.

\subsection{Real-Time Compressed Sensing}

The proposed encoder, though simple in design, proves effective by utilizing a single matrix multiplication, similar to the traditional optimization-based HCS that relies on random projection. This approach draws inspiration from the linear subspace/basis concept in optimization-based HCS, which is known to preserve most of the HSI's energy. This forms the basis of our hypothesis that a linear operation in the decoder could suffice for efficient compressed sensing. Specifically, the encoder in our RTCS system consists of a single convolution operation without any activation function, significantly reducing computational complexity and latency. We employ a stripe-like input to align with the pushbroom scanning mechanism typical in moderate LEOS, as indicated in DCSN \cite{dcsn}. However, our measurement matrix design in compressed sensing diverges from the encoder in DCSN \cite{dcsn}, where a multi-layer approach might lead to considerable encoding delay for hyperspectral stripes and fail to represent a measurement matrix accurately. We contend that single matrix multiplication can act as an efficient encoder, fully meeting the requirements of a measurement matrix in compressed sensing.

Let the original high-resolution (HR) HSI be denoted as $\bX_o\in \mathbb{R}^{B\times H\times W}$, where $B$, $H$, and $W$ represent the number of bands, height, and width of the HRHSI, respectively. The stripe-like HSI, sensed by the pushbroom mechanism of the LEOS, is defined as $
    \bX = \bR \bX_o,
$
where $\bm R$ is the sampling matrix used to acquire the stripe-like HSI $\bX \in \mathbb{R}^{B\times h\times w}$, $h$ and $w$ represent the height and width of the stripe-like HSI. Our compressed sensing in the proposed encoder could be done simply by
\begin{equation}
    \bm Z = \bm \Psi \bm X,
\end{equation}
where $\bm \Psi$ is the measurement matrix formed by convolutional and learnable coefficients. Let us explain in detail how the convolution benefits the HCS. Suppose that a low-dimensional representation $\bZ$ of the HSI could be compressively sensed by a single learnable convolution kernel $W$ via 
\begin{equation}
    Z(m,n) = \sum_{j=-\frac{k_w}{2}}^{\frac{k_w}{2}} \sum_{i=-\frac{k_h}{2}}^{\frac{k_h}{2}} X(i,j) \cdot W(m-i, n-j),
\end{equation}
where $k_w$ and $k_h$ indicate the convolutional kernel's width and height. It is well-known that the convolutional could be expressed as matrix multiplication. Consider a $3\times 3$ data matrix $\bX$ and a $2\times 2$ convolution kernel $\bW$, the convolution operation (stride $s=1$, no padding $p=0$) is equivalent to:

\begin{equation}
\bZ = \bX \ast \bW =
\begin{bmatrix}
x_1 & x_2 & x_3 \\
x_4 & x_5 & x_6 \\
x_7 & x_8 & x_9 \\
\end{bmatrix} \ast
\begin{bmatrix}
w_1 & w_2 \\
w_3 & w_4 \\
\end{bmatrix}.
\end{equation}
This can be reformulated as a matrix multiplication by rearranging $\bW$ with zero-padding and shifting to form $\bW_m$ and vectorizing $\bX$ into its column vector form $\bX_v$:
\begin{equation}
\begin{aligned}
\bZ &= \bW_m\bX_v \\
&=
\setlength{\arraycolsep}{2.5pt} 
\begin{bmatrix}
w_1 & w_2 & 0 & w_3 & w_4 & 0 & 0 & 0 & 0 \\
0 & w_1 & w_2 & 0 & w_3 & w_4 & 0 & 0 & 0 \\
0 & 0 & 0 & w_1 & w_2 & 0 & w_3 & w_4 & 0 \\
0 & 0 & 0 & 0 & w_1 & w_2 & 0 & w_3 & w_4
\end{bmatrix}
\begin{bmatrix}
x_1 \\ x_2 \\ x_3 \\
x_4 \\ x_5 \\ x_6 \\
x_7 \\ x_8 \\ x_9
\end{bmatrix}.
\end{aligned}
\end{equation}
In hyperspectral image compression, exploiting spatial and spectral redundancies is essential. Traditional methods typically reshape 3D hyperspectral data $\bX \in \mathbb{R}^{B\times H\times W}$ into a 2D matrix $\bX_{2D} \in \mathbb{R}^{B\times (HW)}$ to achieve dimensionality reduction. Linear transformations, for example, efficiently capture spectral information and preserve energy, a principle that can be extended to single convolution layers in deep learning frameworks. This strategy is particularly well-suited for the constraints of miniaturized satellite imaging systems, notably those utilizing pushbroom sensing. It effectively reduces the space complexity that is otherwise inherent in compressing the full tensor.

\begin{figure}
    \begin{center}
    \includegraphics[width=0.48\textwidth] {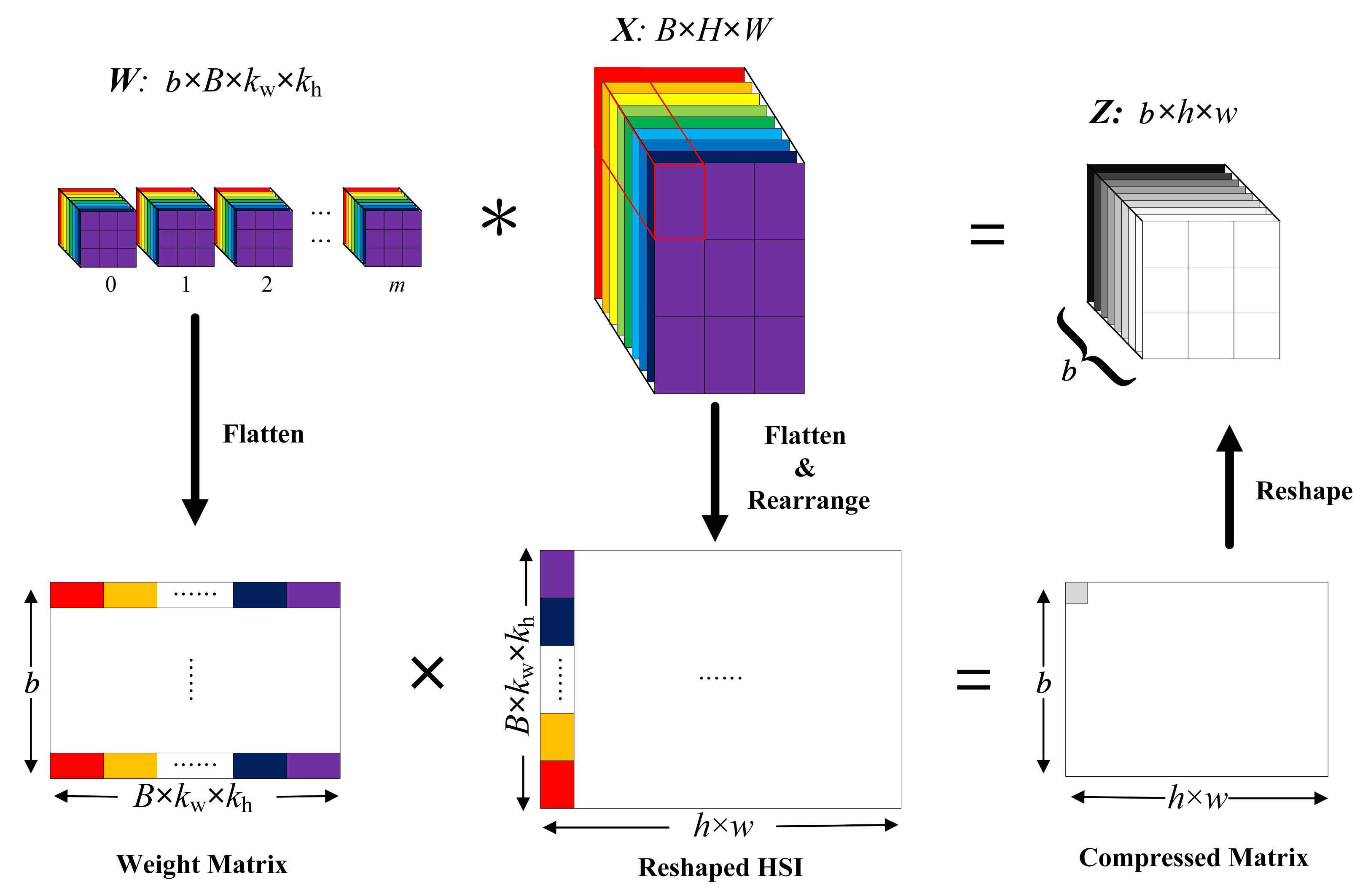}
    \end{center}
    \caption{The diagram of convolutional operation as matrix multiplication for HCS in our RTCS encoder.}
    \label{fig:conv_as_mm}
\end{figure}

A concern is that the typical kernel size $3\times 3$ has a small receptive field in the spatial dimension for HSI. As shown in the convolution example in (3), the compact representation via convolution is determined by only four coefficients $\bm W$, while traditional optimization-based HCS has a larger number of coefficients for dimensionality reduction. Therefore, a larger kernel size, such as $9\times 9$, is introduced to capture the long-range dependency in the spatial domain, resulting in better energy preservation. The kernel size is an example of the effective receptive field. To satisfy the requirements of pushbroom sensing in the miniaturized satellites, we use $9\times 1$ kernel in the encoder to enable the compressed sensing in hyperspectral stripes. Assuming that the input HSI of size $B\times H\times W$ is projected to its compact representation $\bm Z = \bm X \bm \Psi, \in \mathbb{R}^{b\times h\times w}$, where $b \ll B, h \ll H$, and $w \ll W$, $B\times k_h\times k_w$ coefficients are required to project the HSI to a single feature map, and thus, $b\times B\times k_h\times k_w $ coefficients are needed to generate the $b$ feature maps for HSI. The complete matrix multiplication diagram of the convolution operation in the proposed encoder is depicted in Figure \ref{fig:conv_as_mm}, illustrating how convolutional kernels capture the spatial and spectral dependency simultaneously. Specifically, each column vector in the kernel matrix is used to project the row vector of the HSI within a local window determined by kernel size $k_h\times k_w$ to the lower dimensional space, where the spectral redundancy is fully explored by $B\times k_h\times k_w$ coefficients. As the parameter space of the linear convolution is sufficiently large, it should be able to project the HSI to a compact linear subspace.

It is crucial to discuss the computational complexity of the proposed encoder as well. The computational complexity of the proposed encoder, primarily based on matrix multiplication, is $\mathcal{O}(b(Bk_w k_h) (hw))$ with stride $s=1$ and padding $p=0$. In contrast, the major time cost of the most efficient HCS method, AAHCS \cite{aahcsd}, lies in the basis computation via eigendecomposition, where the complexity can be $\mathcal{O}(B^3 + (WH)B(WH))$. In AAHCS \cite{aahcsd}, the vectorized representation $\bX_\text{v}$ of HSI $\bX$ transforms the $B\times H\times W$ matrix into a $B\times (WH)$ matrix and performs eigendecomposition on the square matrix $\bU = \bX_\text{v}\bX_\text{v}^T$ to obtain the basis $\bm \Psi$. Although the optimization-based approach \cite{aahcsd, HYCA} claims that the basis matrix $\bm \Psi$ can be calculated from pre-collected data, their experiments are evaluated by data-dependent basis (i.e., the basis should be calculated for each HSI individually), resulting in eigendecomposition being required onboard the miniaturized satellite during HCS for optimal performance. In comparison to the computational complexity of basis decomposition in AAHCS \cite{aahcsd}, $\mathcal{O}(B^3 + (WH)B(WH))$, the matrix $\bU$ demands $\mathcal{O}((WH)B(WH))$. When $w \ll W$, $h \ll H$, and $b\ll B$, the computational complexity of the proposed encoder can be significantly reduced. For instance, with a sampling rate of $\approx 1\%$ for our encoder, we set the size of compressed HSI to $b=27, h=32, w=1$ (cf. Section \ref{sec:experiments}) for the HSI sized at $172\times 128\times 4$, resulting in computational complexity of approximately $3.95\times 10^6$, while the matrix $\bU$ in AAHSC costs $45.088\times 10^6$. Furthermore, convolutional operations can be highly parallelized in practice, significantly accelerating the computing process and minimizing onboard computational effort in miniaturized satellites. In summary, the proposed encoder is highly efficient and effective.

\subsection{Hardware-Friendly Decoding and Restoration}
\begin{figure}[t]
    \begin{center}
    \includegraphics[width=0.5\textwidth] {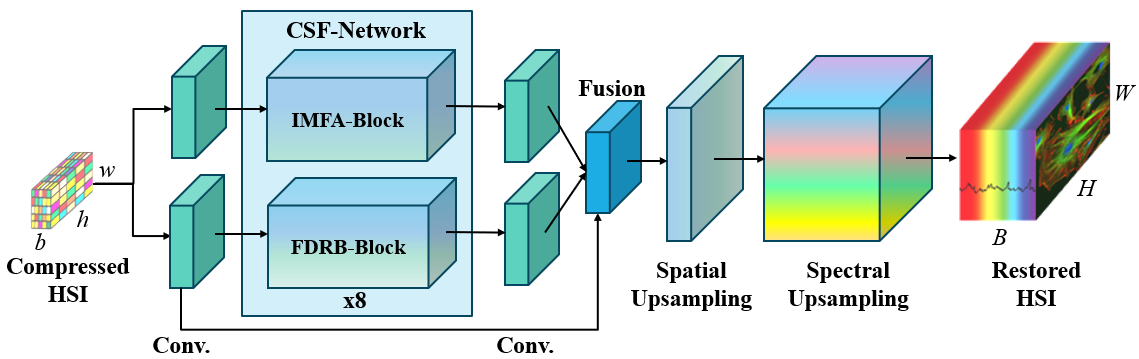}
    \end{center}
    \caption{The overview of the proposed hardware-friendly and effective decoder for HSI compressed sensing and restoration.}
    \label{fig:decoder}
\end{figure}

\begin{figure*}[t]
    \begin{center}
    \includegraphics[width=1\textwidth] 
    {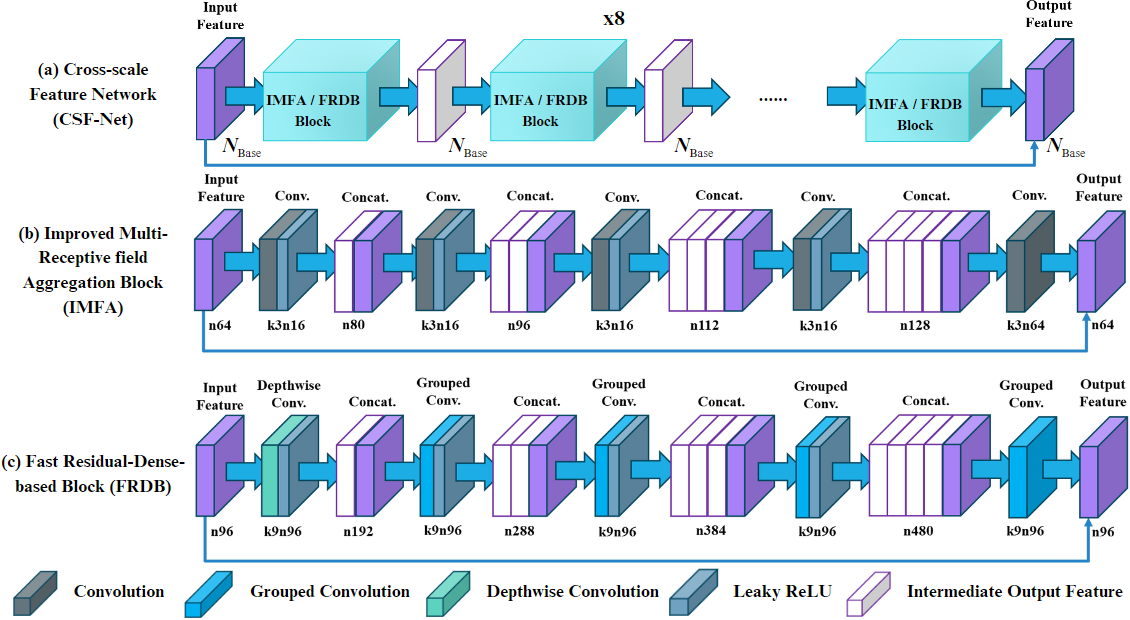}
    
    \end{center}
    \caption{Network architecture of the proposed two-streamed CSF-Net, where k(c)n(c) represents the number of kernels and feature maps respectively. }
    \label{fig:decoder_frdb}
\end{figure*}

\begin{figure}
    \begin{center}
    \includegraphics[width=0.5\textwidth] 
    {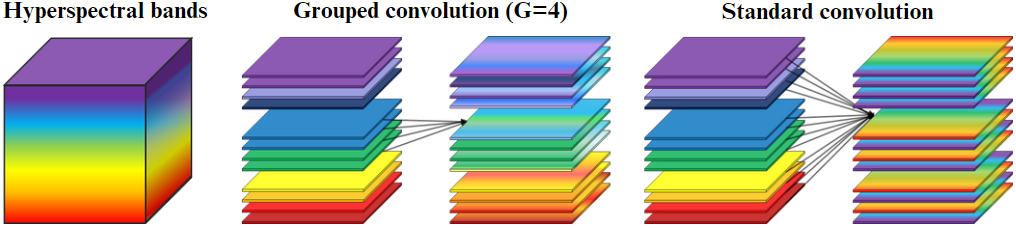}
    \end{center}
    \caption{{{The comparison between grouped and standard convolution lies in their treatment of feature maps for reducing the number of parameters and computational complexity.
}}}
    \label{fig:grouped_conv}
\end{figure}

As previously stated, decoding compressed HSI $\bZ$ in edge devices can effectively reduce the workload of central servers at ground stations. Toward this goal, we propose a novel network structure for efficient decoding of compressed HSI. While reducing the depth and number of kernels in a neural network is an effective way to accelerate inference, overly simplifying the network can detrimentally affect performance, despite achieving faster inference speeds. For instance, in real-time applications, the MobileNet-based architecture \cite{mobilenetv3} is favored due to its computational efficiency. However, there is still a noticeable performance gap between MobileNet and earlier conventional CNN designs like SENet \cite{senet} and ResNet \cite{resnet}. Specifically, MobileNet's best top-1 accuracy on the ImageNet validation set \cite{krizhevsky2012imagenet} is $75.2\%$, compared to $78.57\%$ for ResNet-152 \cite{resnet}. In the context of HCS, particularly during the decoding phase, achieving promising performance is as crucial as ensuring speed.

In this section, we highlight the benefits of our proposed network architecture for the decoder, which is fast, hardware-friendly, and effective. This architecture is illustrated in the Figure \ref{fig:decoder}. Consider the decoder's input as the compressed HSI $\bZ \in \mathbb{R}^{b\times h\times w}$. Our two-stream-based network architecture is designed to learn features with varying receptive fields efficiently. Initially, two stem convolution layers extract low-level features $\bZ_{stem}$ and $\bm \zeta_{stem}$ from $\bm Z$. In one branch, we implement a modified version of the decoder from DCSN \cite{dcsn}, halving its depth and reducing parameters to decrease computational complexity. In the other branch, we propose a Fast Residual-Dense-Block-based Network (FRDB-Net), employing grouped convolution and larger kernels yet maintaining a low parameter size. A Multi-Scale Feature Fusion (MFF) strategy is introduced, combining feature representations from both branches through element-wise addition. Subsequently, using an upsampling block, the low-spatial-resolution (LSR) features are progressively upsampled to high-spatial-resolution (HSR) features. Finally, the HSI is reconstructed from the low-spectral-resolution (LSPR) features using a spectral upsampler via convolution. Detailed descriptions follow in the subsequent paragraphs.\\

\noindent\textbf{Cross-Scale Feature Network.} We first discuss the significant advantages conferred by the two-streamed network architecture in HCS. A direct reduction in network depth can lead to substantial performance loss, as shallower networks often struggle to learn rich feature representations. This is due, in part, to the larger equivalent receptive field of deeper networks compared to shallower ones. Our proposed two-streamed architecture aims to overcome the limitations associated with shallower networks. It comprises two distinct sub-networks utilizing $3\times 3$ and $9 \times 9$ convolutional kernels to capture features with different receptive fields. Feature aggregation from these sub-networks is achieved through element-wise addition using a MFF strategy. This approach allows for sharing cross-scale feature information, which is then progressively upsampled in both spatial and spectral dimensions to reconstruct the HSI. 

The top line of Figure \ref{fig:decoder} presents the first branch, the Improved Multi-Receptive-field Aggregation Network (IMFA-Net), while the bottom line illustrates the second branch, our newly proposed FRDB-Net with a larger kernel size. In developing the IMFA and FRDB blocks, we opted for a single residual connection to minimize memory usage, addressing the issue of feature map reuse seen in DCSN \cite{dcsn}. Our experiments demonstrate that this streamlined approach with just one residual connection is both more effective and efficient.

Specifically, the intermediate feature $\bm Z_\text{m}$ could be extracted via
\begin{equation}
    \bm Z_m = f_\text{u}(f_\text{IMFA}(f_{\text{s}_1}(\bZ))) + f_\text{FRDB}(f_{\text{s}_2}(\bZ)),
\end{equation}
where $f_\text{FRDB}$ and $f_\text{IMFA}$ are the backbone networks of the reduced IMFA and the proposed FRDB, $f_{\text{s}_1}$ and $f_{\text{s}_2}$ are the stem convolution layers for IMFA and FRDB. In IMFA-Net and the proposed FRDB-Net, we stack $n_f$ IMFA and FRDB blocks, respectively, where $n_f=8$ in this study. Since the number of the feature maps of $f_\text{IMFA}$ is less than that of $f_\text{FRDB}$, an extra convolution layer $f_\text{u}$ is used to project the feature map to the desired size. Then, the HSR feature map $\bm Z_h$ could be learned based on the spatial upsampler via
\begin{equation}\label{eq:upsample}
    \bm Z_\text{h} = f_\text{up}(\text{Bilinear}(\bm Z_\text{m})),
\end{equation}
where Bilinear stands for the bilinear interpolation function and $f_\text{up}$ is a $3\times 3$ convolution layer. By repeating the function \ref{eq:upsample}, the feature map with higher spatial resolution could be obtained. Finally, the high spectral resolution HSI could be recovered from the HSR feature map via
\begin{equation}
    \bm X^* = f_\text{rec}(\bm Z_\text{h}),
\end{equation}
where $f_\text{rec}$ is the convolution layer for mapping 96-channeled feature maps to $B$-channeled ones. Note that the kernel size of most convolution layers is $3\times 3$, except for the proposed FRDB-Net. The rest of the question is the design of the reduced IMFA and the proposed FRDB networks. More specifically, a larger kernel size used in the proposed FRDB could result in heavy computational effort, how to effectively reduce the complexity of the FRDB is essential.\\

\noindent\textbf{Lightweight Design.} In this subsection, we introduce the base blocks for IMFA-Net and FRDB-Net, which are both effective and efficient. The detailed network architectures of these blocks are depicted in Figure \ref{fig:decoder_frdb}. As illustrated in Figure \ref{fig:decoder}, each block in both IMFA-Net and FRDB-Net comprises three stacked base blocks with a shortcut connection linking the input and output feature maps (Figure \ref{fig:decoder}(a)). Within each base block, the input feature map is condensed into a $c_s$-channeled feature map and merged with the input feature map, resulting in a $c_s + N_\text{Base}$-channeled feature map (Figure \ref{fig:decoder_frdb}(b)(c)). Following DCSN \cite{dcsn}, we set $N_\text{Base}$ to 64 and reduce $c_s$ to 16 from 32, decreasing computational complexity.

For the FRDB-Net, we draw inspiration from \cite{dcsn, sr2} and use a residual dense block (RDB) to construct its base block. We identify that feature map reuse is the primary complexity factor in the RDB feed-forward process. Utilizing depthwise separable convolution, as in MobileNet, effectively reduces parameters and computational complexity. Additionally, grouped convolution, employed in ResNeSt, further decreases complexity with an optimal number of groups $c_g$. We apply grouped convolution in our FRDB base block to achieve a balance between parameter count and complexity. Specifically, we set $c_g=N_\text{Base}=c_s=96$, reducing the computational complexity in the last layer of the FRDB base block (Figure \ref{fig:decoder_frdb}(b)) by a factor of 5. This approach shows that grouped convolution can lower computational complexity without compromising performance.

The 2D convolution exploits spatial domain local correlation for complexity reduction, just as the high redundancy between HSI bands permits further computational efficiency. Figure \ref{fig:grouped_conv} illustrates the efficacy of grouped convolution in this context. Given the higher redundancy among successive HSI bands than distant ones, grouped convolution naturally leverages this redundancy, significantly reducing computational complexity and parameter count without sacrificing performance. For example, with a $172\times 128\times 4$ stripe-like HSI, one feature map can be derived from 43 bands (172/ 4) using grouped convolution when $c_g=4$, efficiently exploring local redundancy in the spectral domain. The proposed decoder, incorporating IMFA-Net, capitalizes on cross-scale feature interaction, ensuring promising performance. 

\subsection{Task-Specific Optimization}
To boost efficiency, especially in the grouped convolution of our proposed FRDB-Net, we have significantly reduced the number of channels in the feature maps of the RTCS. This reduction could potentially lead to a slight decrease in performance, particularly in terms of spectral quality, due to the constrained parameter space. For instance, the proposed method comprises approximately $6.291M$ parameters, in contrast to the $11.951M$ parameters in DCSN \cite{dcsn}. DCSN \cite{dcsn} employs solely the $\ell_1$ loss for training, which might not fully address the variance across different bands, reflecting diverse values. This could mean that the $\ell_1$ loss might inadvertently prioritize minimizing bands with higher values, potentially compromising spectral fidelity in certain bands.

Recognizing the paramount importance of spectral index in HCS, we introduce a SAM loss function to guide the proposed RTCS towards superior spectral index outcomes more effectively. The SAM loss is defined as:
    \begin{equation}
    \ell_{\text{SAM}} = \frac{1}{N} \sum_{n=1}^{N} \cos^{-1} \left( \text{clip} \left( \frac{\bm{X}_n^T \bm{X}^*_n + \epsilon}{\|\bm{X}_n\|_2 \cdot \|\bm{X}_n^*\|_2 + \epsilon} \right) \right)
    \end{equation}
where $\bm{X}_n$ and $\bm{X}^*_n$ represent the $n$-th spectral vector of the ground truth and the reconstructed HSI, respectively. The SAM loss calculates the angular difference between these spectral vectors, using the arc-cosine of the normalized dot product to measure cosine similarity. We add a small constant $\epsilon$ to the denominator to prevent division by zero, thus enhancing numerical stability. The clip function constrains the similarity range between -1 and 1, ensuring stable arc-cosine calculations. By focusing on angular differences instead of magnitudes, this loss provides a robust metric for assessing the spectral fidelity of the reconstructed HSI compared to the original. It emphasizes the orientation of spectral signatures over their amplitude, facilitating a more accurate comparison of spectral characteristics.

\begin{equation}
    \ell_\text{T}(\bm X, \bm X^*) = \lVert \bm X - \bm X^* \rVert + \alpha \ell_\text{SAM}(\bm X, \bm X^*),
\end{equation}
where $\alpha$ is the weighting factor to control the balance between $\ell_1$ and the proposed SAM loss. 
Furthermore, the stripe effect is often presented in the existing hyperspectral image datasets due to sensor errors or electrical exceptions. We also introduce this phenomenon into the training process via random masking on the hyperspectral stripes. In this fashion, we could enable the proposed RTCS to restore the complete hyperspectral stripes from the missing one during the decoding stage. Specifically, let the binary mask be $\bm M \in \mathbb{R}^{B\times H\times W}$, the masked HSI could be
\begin{equation}
    \bm X_\text{mask} = \bm X \odot \bm M.
\end{equation}

In our proposed method, the use of a mask is crucial, yet its design remains relatively simple. The diverse characteristics of the stripe effect make it impractical to capture all possible types during the training process. To address this, we have implemented a basic mask design that involves randomly masking parts of successive bands within a set of hyperspectral stripes. Based on our research, we found that the likelihood of a hyperspectral stripe being affected by stripe effects is about 0.2, and the maximum number of consecutive bands impacted by stripe effects is also 0.2. This strategic approach effectively equips the proposed RTCS technique to reconstruct hyperspectral stripes while reducing the impact of the stripe effect. It ensures that the RTCS can effectively reconstruct hyperspectral data using only partially available compressed information.

In order to guide the proposed RTCS in effectively restoring the HSI with stripe effect, the total loss function used for data imputation could be

\begin{equation}
\begin{aligned}
    \ell_\text{Aug.} &= \ell_\text{T}(\bm X, \bm X^*) \\
    &+ \lVert \bm X_\text{mask} - \bm X^*_\text{mask} \rVert + \alpha \ell_\text{SAM}(\bm X_\text{mask}, \bm X^*_\text{mask}).
\end{aligned}
\end{equation}
Finally, the robust RTCS could be trained with the proposed total loss function $ \ell_\text{Aug.}$. The mask selection and design used in $\ell_\text{Aug.}$ are discussed in the experiments. 
		
	\section{Experimental Results}\label{sec:experiments}

\subsection{Experimental Settings}\label{sec:settings}

	\noindent\textbf{Dataset descriptions.} 
    The dataset used in this study was collected from AVIRIS \cite{AVIRISrealdata} sensor, which was pre-processed in the same way as DCSN \cite{dcsn}. We collected the HSI data between 2008-2012, and the original HSI data was partitioned into 256 $\times$ 256 sub-images, resulting in 1735 HSI sub-images. We then applied the k-means to cluster the HSI sub-images into five types, including \textbf{C}ity, \textbf{M}ountain, \textbf{F}orest, \textbf{F}arm, and \textbf{O}ther types, and manually corrected any misclassified HSI sub-images. We selected 200 HSI sub-images from each class, resulting in a total of 1200 HSI sub-images as the dataset. The dataset was split into training, validation, and testing sets, consisting of 1000, 100, and 100 HSI sub-images. The dataset will be released.

    The size of the stripe-like HSI is designed to be $172\times 128 \times 4$ in the training phase, while $172\times 256\times 4$ stripe-like HSI is used in the evaluation phase, leading to that only very limited memory usage is required for storing four hyperspectral stripes for compressed sensing. But the proposed RTCS still works well for a single hyperspectral stripe-based compressed sensing. To evaluate the performance, we stack the decoded HSI stripes to restore the HSI sub-image with $172\times 256\times 256$ since the optimization-based approach requires full 3D-tensor as the input, followed by measuring their objective quality by the metrics in \cite{dcsn}, including Peak signal-to-noise ratio (PSNR), Root-mean-square error (RMSE), and the Spectral Angle Mapper (SAM).

	\noindent\textbf{Hyper-parameter settings.}   
    For the training phase of our RTCS model, the AdamW optimizer \cite{adamw,loshchilov2018decoupled}, an improved version of traditional Stochastic Gradient Descent (SGD) algorithm, is employed to train our network, where ($\beta_1$, $\beta_2$) are (0.9, 0.999), and the epsilon $\epsilon$ is $1\textrm{e}\!-\!8$. Considering the possible diversity of HSI datasets, including exhibiting extremely high variability due to their rich spectral information and complex spatial features, the AdamW optimizer is crucial to enhance the generalization ability of the model.
    The initial learning rate is $1\textrm{e}\!-\!4$ with the step-wise learning rate decaying every 1,000 epochs with scaling factor $1/2$, and the batch size is set to $10$. The number of total epochs of the proposed RTCS in the training phase is 5,000. The sampling rate $s_r$ of the proposed RTCS is determined by the size of the compact representation $\bm Z$ of the HSI $\bm X$. Specifically, the spatial size of the HSI is reduced to be $\frac{1}{16}$ (i.e., $\frac{1}{4}$ for width and height) for the target sampling rate $s_r \leq 1\%$, while the number of the feature maps $b$ of the compressed HSI $\bm Z$ could be calculated based on $b =\lfloor  B\times s_r \times 16 \rfloor$. While the sampling rate $s_r\geq 1\%$, the spatial reduction is $\frac{1}{4}$ and the number of channels of $\bm Z$ could be $b =\lfloor  B\times s_r \times 4 \rfloor$.

				
    \begin{table*}[t]
	\vspace{0.1cm}
	\centering 
	\caption{Performance comparison between the proposed RTCS and other HCS methods under different sampling rates (SR) $\approx 0.5\%$, $1\%$, and $5\%$.  The result in \textbf{bold} denotes the best method under the same sampling rate. Star (*) denotes the method failed at SR=$0.5\%$. The data marked with *SS are evaluation metrics that use only single stripe data.}
	\scalebox{0.75}{
		\begin{tabular}{|l|c| c c c c c |c|}\hline
			\multirow{2.2}{*}{Method} 
			  & \multirow{2.2}{*}{SR} & \multicolumn{6}{c|}{{{Test set}} (PSNR$\uparrow$ / RMSE$\downarrow$ / SAM$\downarrow$)} \\ \cline{3-8}
			 & & C-type & M-type &	F-type & L-type & O-type & Averaged \\ \hline
            \multirow{2.2}{*}{AAHCS*\cite{aahcsd}} & $1\%$ & 31.295 / 220.932 / 7.738 & 33.247 / 65.114 / 3.609 & 31.611 / 91.802 / 4.809 & 31.337 / 115.610 / 3.920 & 31.611 / 91.802 / 4.809 & 32.978 / 107.917 / 4.287 \\ \cline{2-8}
            & $5\%$ & 43.400 / 66.755 / 2.171 & 44.056 / 23.598 / 1.215 & 44.403 / 23.854 / 1.103 & 44.286 / 33.319 / 1.033 & \textbf{44.403} / \textbf{23.854} / 1.103 & 44.784 / 32.820 / 1.202 \\ \hline
            
            \multirow{2.2}{*}{SPACE*\cite{space}} & $1\%$ & 26.275 / 437.212 / 8.861 & 22.073 / 389.708 / 6.077 & 22.389 / 446.935 / 8.131 & 21.340 / 879.260 / 6.902 & 29.445 / 245.843 / 2.441 & 24.304 / 479.792 / 6.483 \\ \cline{2-8}
            & $5\%$ & 38.607 / 56.614 / 2.171 & 38.021 / 26.525 / 1.597 & 37.511 / 27.172 / 1.485 & 37.901 / 34.258 / 1.500 & 39.092 / 27.860 / 0.898 & 38.226 / 34.486 / 1.530 \\ \hline

            \multirow{2.2}{*}{SpeCA*\cite{martin2016hyperspectral}} & $1\%$ & 15.092 / 590.184 / 22.785 & 19.122 / 216.356 / 13.100 & 13.867 / 421.521 / 19.745 & 12.814 / 535.711 / 22.127 & 23.255 / 168.216 / 7.714 & 16.830 / 386.398 / 17.094 \\ \cline{2-8}
            & $5\%$ & 27.942 / 180.709 / 6.627 & 29.225 / 68.090 / 3.873 & 27.615 / 83.614 / 4.178 & 27.424 / 117.731 / 4.795 & 32.417 / 67.002 / 1.800 & 28.924 / 103.429 / 4.255 \\ \hline

            & $0.5\%$ & 14.854 / 429.016 / 12.990 & 15.929 / 213.824 / 11.357 & 16.182 / 286.819 / 13.775 & 16.824 / 301.670 / 12.180 & 19.161 / 218.439 / 8.197 & 16.590 / 289.954 / 11.700 \\ \cline{2-8}
            TenTV \cite{wang2017compressive} & $1\%$ & 16.921 / 356.217 / 10.692 & 17.612 / 169.388 / 8.749 & 18.195 / 210.931 / 9.709 & 18.891 / 220.237 / 8.822 & 19.484 / 164.420 / 5.856 & 18.220 / 224.239 / 8.766 \\ \cline{2-8}
            & $5\%$ & 22.553 / 208.414 / 6.793 & 23.177 / 97.187 / 5.101 & 24.149 / 108.601 / 4.867 & 24.924 / 111.323 / 4.411 & 25.313 / 84.591 / 2.916 & 24.023 / 122.023 / 4.818 \\ \hline
  
            & $0.5\%$ & 21.885 / 1949.291 / 12.710 & 18.691 / 338.328 / 12.089 & 18.361 / 359.933 / 11.898 & 16.375 / 586.444 / 14.185 & 21.779 / 387.668 / 7.313 & 19.418 / 724.333 / 11.639 \\ \cline{2-8}
            H-LSS\cite{zhang2016locally} & $1\%$ & 24.120 / 324.823 / 8.691 & 21.803 / 244.465 / 9.023 & 22.491 / 197.237 / 8.262 & 20.687 / 329.950 / 9.498 & 25.405 / 229.222 / 4.799 & 22.901 / 265.139 / 8.054 \\ \cline{2-8}
            & $5\%$ & 27.622 / 194.043 / 6.247 & 28.720 / 79.231 / 4.730 & 26.769 / 104.582 / 5.315 & 26.292 / 129.837 / 5.716 & 32.007 / 70.704 / 2.574 & 28.282 / 115.679 / 4.916 \\ \hline
 
            & $0.5\%$ & 29.398 / 152.900 / 3.733 & 30.588 / 59.821 / 2.466 & 30.674 / 69.437 / 2.454 & 31.340 / 83.893 / 2.465 & 32.714 / 54.669 / 1.357 & 30.943 / 84.144 / 2.495 \\ \cline{2-8}
            DCSN\cite{dcsn} & $1\%$ & 31.446 / 127.919 / 2.745 & 32.450 / 48.979 / 1.878 & 32.372 / 58.523 / 1.897 & 32.986 / 69.670 / 1.879 & 34.263 / 47.262 / 1.064 & 32.703 / 70.471 / 1.893  \\ \cline{2-8}
            & $5\%$ & 35.938 / 86.090 / 1.101 & 37.651 / 30.632 / 0.797 & 36.873 / 39.761 / 0.832 & 37.251 / 47.666 / 0.757 & 38.726 / 31.610 / 0.467 & 37.288 / 47.152 / 0.791 \\ \hline
   
            \hhline{|=|=======|}
            & $0.5\%$ & \textbf{31.601} / \textbf{120.754} / \textbf{2.377} & \textbf{32.034} / \textbf{50.676} / \textbf{1.680} & \textbf{32.583} / \textbf{54.021} / \textbf{1.590} & \textbf{33.617} / \textbf{61.818} / \textbf{1.415} & \textbf{34.862} / \textbf{44.466} / \textbf{0.867} & \textbf{32.939} / \textbf{66.353} / \textbf{1.586} \\ \cline{2-8}
            RTCS & $1\%$ & \textbf{36.036} / \textbf{74.521} / \textbf{1.66} & \textbf{36.113} / \textbf{31.814} / \textbf{1.215} & \textbf{36.129} / \textbf{33.908}  / \textbf{1.169} & \textbf{36.952} / \textbf{40.374} / \textbf{1.092} & \textbf{37.455} / \textbf{37.788} / \textbf{0.734} & \textbf{36.537} / \textbf{43.681} / \textbf{1.176} \\ \cline{2-8}
            & $5\%$ & \textbf{47.795 / 17.187 / 0.608} & \textbf{45.535 / 9.571 / 0.578} & \textbf{44.479 / 11.840 / 0.611} & \textbf{44.367 / 16.573 / 0.620} & 43.376 / 22.100 / \textbf{0.497} & \textbf{45.110 / 15.455 / 0.583} \\ \hline
            
            RTCS *SS & $1\%$ & 36.767 / 67.353 / 1.754 & 35.571 / 32.479 / 1.352 & 35.797 / 33.791 / 1.274 & 37.119 / 38.141 / 1.178 & 37.894 / 31.161 / 0.784 & 36.629 / 40.585 / 1.269 \\ \cline{2-8} \hline
	\end{tabular}}
	\vspace{0.1cm}
	\label{tab:cmp_rtcsCR}
    \end{table*}
    
\subsection{Performance Evaluation}\label{sec:performance_eval}
    To illustrate the superiority of RTCS, seven state-of-the-art HCS methods are selected as baselines, including AAHCS\cite{aahcsd}, SPACE\cite{space}, SpeCA\cite{martin2016hyperspectral}, TenTV\cite{wang2017compressive}, Hyper-LSS (H-LSS) \cite{zhang2016locally}, and the prior art, DCSN\cite{dcsn}. 

    We follow the standard HSI-based metrics used in DCSN \cite{dcsn}, including PSNR, RMSE, and SAM, to evaluate the performance of the proposed and other peer methods, where PSNR and RMSE are used to evaluate the spatial fidelity while SAM is adopted to verify the degree of spectral preservation of the reconstructed HSI. 

\subsubsection{\textbf{Objective Quality Assessment}} \label{sec:objective_quality_assessment} 
To demonstrate the robustness and generalizability of our RTCS framework, we carried out cross-type dataset evaluations. The RTCS was trained on one type from the dataset and then tested across four different types. As shown in Table \ref{tab:cmp_rtcsCR}, the performance of RTCS is benchmarked against other methods using metrics like SAM, PSNR, and RMSE. Across all these indices, RTCS consistently surpasses other methods, proving its robustness even when applied to a single hyperspectral stripe without relying on a large memory buffer, as evidenced by the last row in the Table \ref{tab:cmp_rtcsCR}.

Significantly, RTCS shows superior performance compared to other deep-learning-based HCS methods, such as DCSN \cite{dcsn}, despite being trained on a smaller dataset. While DCSN uses 2,300 training samples, RTCS requires only 1,000, highlighting the efficiency of our model in a limited-data scenario. This difference indicates that more complex network structures might need larger datasets for similar performance levels. By simplifying the network architecture, RTCS not only accommodates but excels in few-shot learning scenarios, demonstrating its practicality and efficiency.

Furthermore, as the sampling rate increases, for instance, to $5\%$, most HCS methods, including RTCS, AAHCS \cite{aahcsd}, and DCSN \cite{dcsn}, show significant performance improvements, as seen in the Table \ref{tab:cmp_rtcsCR}. In AAHCS , self-similarity regularization in the spatial domain effectively aids the restoration task, yielding comparable PSNR and RMSE for the reconstructed HSI. However, RTCS outperforms AAHCS in terms of SAM, suggesting superior preservation of spectral signatures by RTCS.

    Consider a scenario where bandwidth is severely limited due to an increase in the number of miniaturized satellites. In such cases, a viable solution might involve reducing the sampling rate of the proposed RTCS even further. As shown in the Table \ref{tab:cmp_rtcsCR}, the RTCS demonstrates considerable performance at an approximate sampling rate of 0.5\%, while the performance of DCSN \cite{dcsn} markedly declines at such low rates. It is worth noting that some partial optimization-based HCS methods, such as AAHCS \cite{aahcsd} and SPACE \cite{space}, couldn't assess performance at this low sampling rate due to significant memory usage and prolonged decoding times. Consequently, the proposed RTCS stands out, especially at lower sampling rates ($s_r$), achieving state-of-the-art performance compared to learning-based HCS methods like DCSN. Furthermore, optimization-based HCS methods generally struggle to deliver satisfactory results at very low sampling rates.

    When comparing the performance of the proposed RTCS with other advanced HCS methods, RTCS distinctly surpasses them. Notably, RTCS effectively manages the stripe effect and transmission noise associated with pushbroom scanning, delivering promising results. Additionally, RTCS is fast, lightweight, and hardware-friendly, making it suitable for real-time applications without compromising its leading performance. In Section \ref{sec:robust}, we will delve into the robustness of the proposed RTCS, encompassing aspects such as restoration with stripe effect, generalizability, and resistance to transmission noise.
    
    

    
\subsection{Robustness Evaluation}\label{sec:robust} 
While the performance of the proposed RTCS achieves state-of-the-art in the ideal case, the robustness and stability evaluation of the proposed RTCS for real-world scenarios of the HCS and data transmission could be more essential. We conduct the robustness evaluation in this subsection, including the cases of channel noise during transmission, stripe effect due to partially and temporarily broken sensors, and few-shot compatibility.

\begin{figure}
    \begin{center}
    \includegraphics[width=0.5\textwidth] {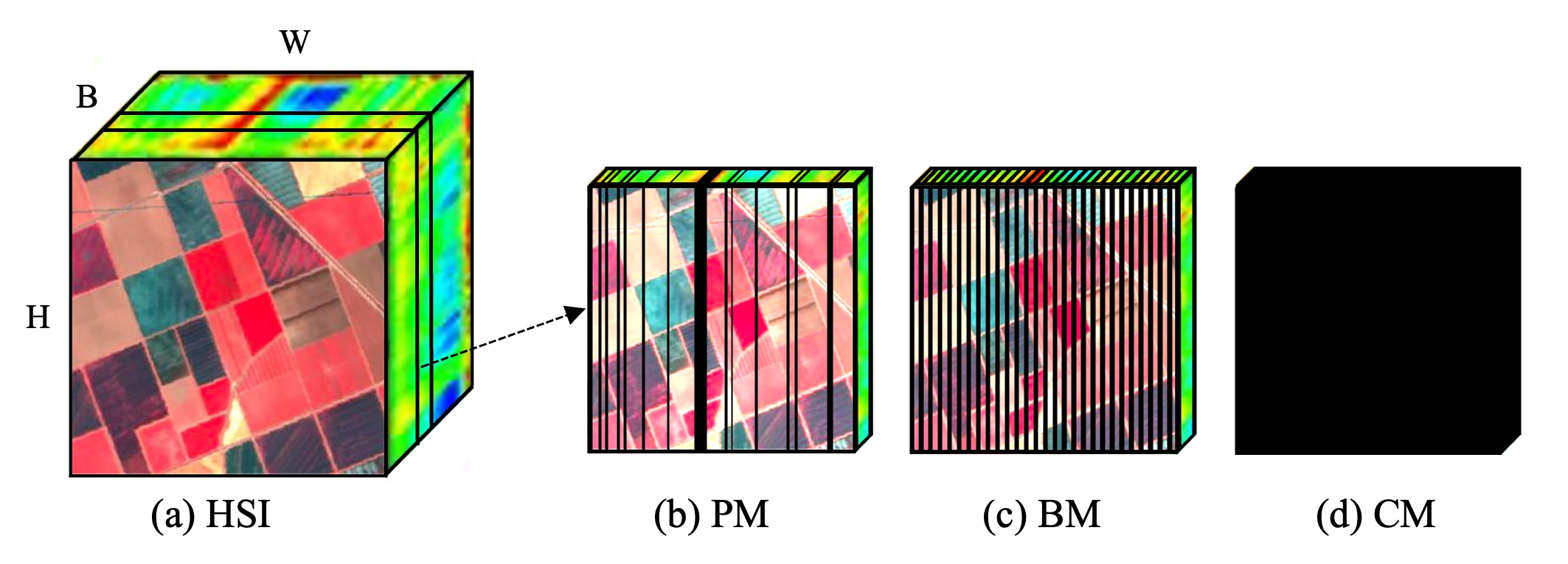}
    \end{center}
    \caption{The visualized examples of (a) hyperspectral image (HS) for three missing types, including (b) partially missing (PM), (c) band-wise missing (BM), and (d) completely missing (CM).}
    \label{fig:missing}
\end{figure}

\subsubsection{\textbf{Restoration with Missing Data}}
As longer-running sensors at the miniaturized satellite in space, a partial spectrum could be missing during compressed sensing, known as the stripe effect of HSI. Most approach treats this could be resolved by post-processing, known as HSI inpainting techniques \cite{admmadam}. With the currently accessible data portal of HSI, some stripe-type stripe effects in the multiple successive bands could be found in \cite{admmadam}, which is also shown in the HSI inpainting approach ADMM-ADAM \cite{admmadam}). In this study, we show that the proposed RTCS can recover the complete HSI from the corrupted HSI using the proposed loss function $\ell_\text{Aug.}$. 

To evaluate the robustness of the stripe effect, three types of stripe effect of HSI are studied in this experiment, as follows:
\begin{itemize}
 
    \item PM (Partially missing): Some of the lines of the HSI stripe are sometimes missing at continuous bands due to unexpected electrical errors on the miniaturized satellite, as an example shown in the Figure \ref{fig:missing}(a). It is the lightest missing type.
    \item BM (Band-wise missing): The sensor for a range of the bands is partially broken, resulting in partially corrupted data in each HSI stripe, as shown in the regular pattern of Figure \ref{fig:missing}(b).
    \item CM (Completely missing): The sensor for a range of the bands is completely broken, resulting in no information in the continuous bands, as shown in the Figure \ref{fig:missing} (c).
\end{itemize}
Only PM and CM are adopted for $\ell_\text{Aug.}$ with the randomly selected stripes and the continuous bands in the training phase. In the testing phase, the fixed missing pattern for three types is generated by the fixed random seed to ensure the fairness of the performance comparison between the proposed and other HCS methods. Note that the BM is not involved in the training phase; thus the results of the BM in the testing phase could be regarded as the generalizability evaluation. 

    \begin{table*}[t]
	\vspace{0.2cm}
	\centering 
	\caption{Robustness evaluation of the different stripe effect types for the proposed RTCS and other peer methods. The result in \textbf{bold} denotes the best method under the same setting of missing bands. (sampling rate $\approx 1\%$)}
	\scalebox{1}{
		\begin{tabular}{|l|c| c c c | c|}\hline
			\multirow{2.2}{*}{Method} 
			  & \multirow{2.2}{*}{{{Missing bands}}} & \multicolumn{4}{c|}{Missing type (PSNR$\uparrow$ / RMSE$\downarrow$ / SAM$\downarrow$)} \\ \cline{3-6}
			 & & {{Band-wise missing}} & {{Partially missing}} & {{Completely missing}} & Averaged \\ \hline
			 
            \multirow{2.2}{*}{AAHCS\cite{aahcsd}} & 50-60 & 31.547 / 1138.582 / 7.091 & 31.613 / 1158.320 / 6.174 & 31.666 / 1133.931 / 4.309 & 31.608 / 1143.611 / 5.858\\ \cline{2-6}
            & 50-80 & 28.639 / 1107.729 / 9.758 & 29.146 / 1147.370 / 7.674 & 29.000 / 1079.546 / 4.267 & 28.928 / 1111.548 / 7.233 \\ \hline
            
            \multirow{2.2}{*}{SpeCA\cite{martin2016hyperspectral}} & 50-60 & 14.610 / 445.688 / 20.848 & 15.196 / 428.369 / 18.752 & 14.880 / 447.976 / 21.431 & 14.895 / 440.667 / 20.343 \\ \cline{2-6}
            & 50-80 & 11.565 / 563.434 / 25.992 & 12.751 / 510.790 / 22.620 & 10.945 / 616.859 / 29.180 & 11.753 / 563.694 / 25.907 \\ \hline

            \multirow{2.2}{*}{TenTV\cite{wang2017compressive}} & 50-60 & 17.784 / 274.052 / 11.458 & 18.199 / 228.665 / 8.955 & 16.988 / 325.028 / 14.194 & 17.657 / 275.915 / 11.535\\ \cline{2-6}
            & 50-80 & 16.802 / 338.644 / 14.992 & 17.798 / 244.089 / 9.713 & 13.946 / 489.591 / 23.134 & 16.182 / 357.441 / 15.946 \\ \hline
            
            \multirow{2.2}{*}{H-LSS\cite{zhang2016locally}} & 50-60 & 20.719 / 423.186 / 14.204 & 21.860 / 384.037 / 8.756 & 22.158 / 403.553 / 16.952 & 21.579 / 403.592 / 13.304 \\ \cline{2-6}
            & 50-80 & 19.553 / 433.793 / 17.562 & 21.153 / 388.866 / 9.073 & 20.434 / 509.740 / 24.233 & 20.373 / 444.133 / 16.956 \\ \hline

            \multirow{2.2}{*}{SPACE\cite{space}} & 50-60 & 26.69 / 243.598 / 11.740 & 28.546 / 135.776 / 5.634 & 24.899 / 298.634 / 14.563 & 26.711 / 226.002 / 10.645 \\ \cline{2-6}
            & 50-80 & 22.540 / 320.169 / 15.180 & 26.119 / 172.501 / 6.638 & 17.632 / 452.930 / 22.559 & 23.430 / 315.2 / 14.792\\ \hline

            \multirow{2.2}{*}{DCSN\cite{dcsn}} & 50-60 & 26.562 / 133.927 / 4.543 & 28.681 / 102.926 / 3.173 & 25.443 / 149.647 / 5.379 & 26.895 / 128.833 / 4.365 \\ \cline{2-6}
            & 50-80 & 22.169 / 216.315 / 8.867 & 26.110 / 139.823 / 4.447 & 19.314 / 288.096 / 12.701 & 22.531 / 214.744 / 8.671\\ \hline
            
            \hhline{|=|====|=}
            
            \multirow{2.2}{*}{RTCS} & 50-60 & \textbf{33.628 / 57.824 / 1.798} & \textbf{35.261 / 48.993 / 1.396} & \textbf{33.318 / 59.726 / 1.881} & \textbf{34.069 / 55.514 / 1.691} \\ \cline{2-6}
            & 50-80 & \textbf{32.450 / 64.309 / 2.079} & \textbf{35.157 / 49.389 / 1.408} & \textbf{33.058 / 61.972 / 2.002} & \textbf{33.555 / 58.556 / 1.829} \\ \hline
	\end{tabular}}
	\vspace{0.2cm}
	\label{tab:inpaint_exp}
    \end{table*}
    
\begin{figure*}
    \begin{center}
    \includegraphics[width=1\textwidth] {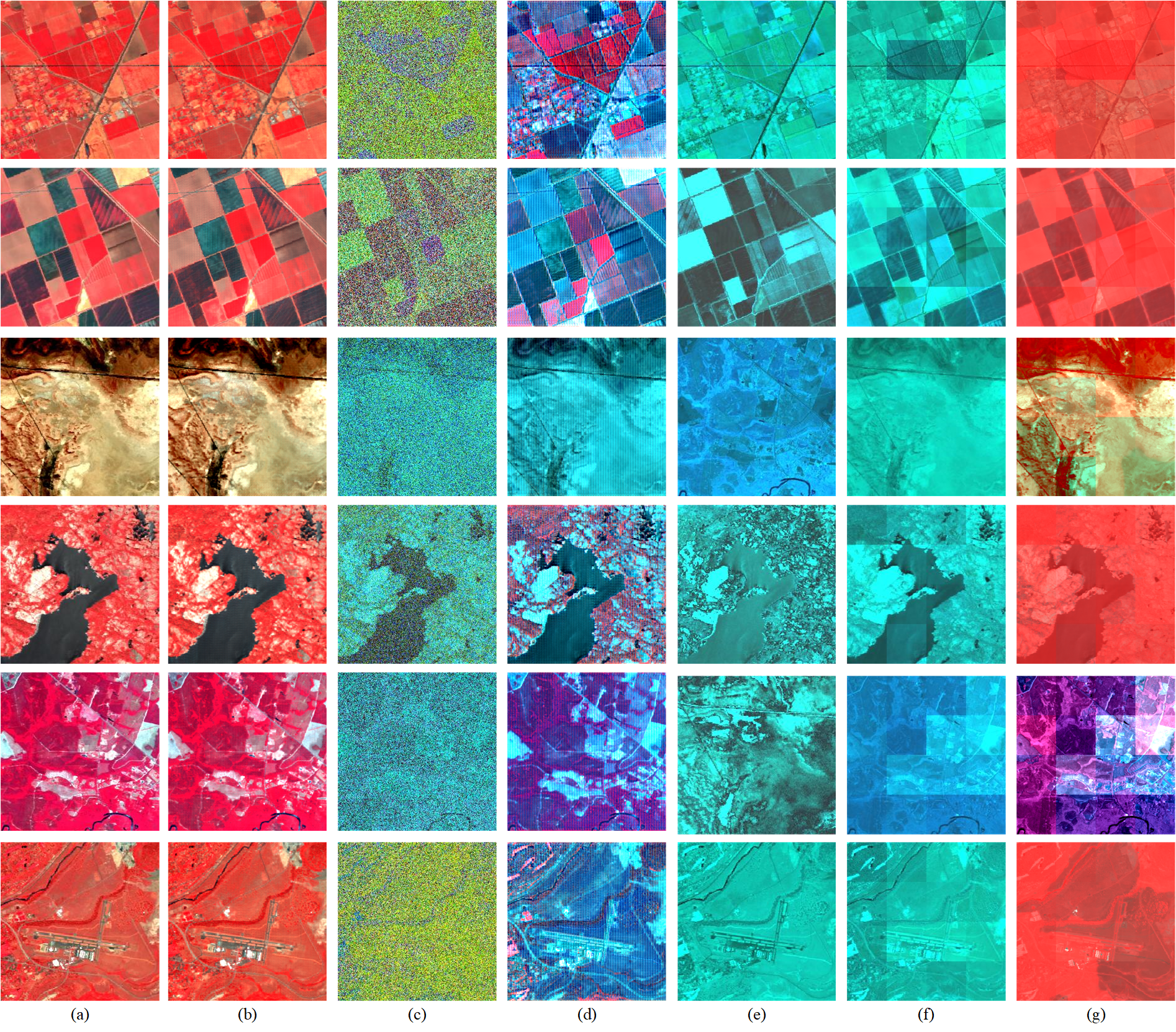}
    \end{center}
    \caption{The false color representation of the visualized hyperspectral image of the (a) ground truth, and the reconstructed counterpart under completely missing (CM) bands from 50-80 using (b) the proposed RTCS, (c) SpeCA \cite{martin2016hyperspectral}, (d) DCSN \cite{dcsn}, (e) H-LSS \cite{martin2016hyperspectral}, (f) TenTV \cite{wang2017compressive} , and (g) AAHCS \cite{aahcsd} and followed by inpainting the stripe effect using ADMM-ADAM \cite{admmadam}.}
    \label{fig:inpainting}
\end{figure*}

Table \ref{tab:inpaint_exp} presents the performance comparison between the proposed RTCS and other HCS methods for three stripe effect types. Note that the sampling rate $s_r$ in the experiments is trained and evaluated on $\approx 1\%$. All the stripe effects could result in significant performance degradation for most of the HCS methods, except for the proposed RTCS. An even worse case is assumed that the partial sensors for the specific bands are completely broken (i.e., CM), resulting in a discontinuity at a range in the spectral domain. In this case, the spectral signature (i.e., SAM) of the reconstructed HSI is significantly degraded for most of the HCS methods under the stripe effect. However, the proposed RTCS still remains strong, even in successive 30 bands completely missing. In BM type, the proposed RTCS also presents promising performance, implying that the proposed RTCS has the ability to deal with the new missing types which are not observed in the training phase. While the DCSN achieves promising performance on the pure HCS, the stripe effect still significantly harms the effectiveness due to the lack of the proposed $\ell_\text{Aug}$. In this way, the compressed hyperspectral stripes could be well reconstructed based on the spectral and spatial context, greatly reducing the stripe effect of the reconstructed HSI based on the decoded and restored hyperspectral stripes. Figure \ref{fig:inpainting} reveals the visualized results using the false color of the reconstructed HSI based on the proposed RTCS and other peer methods in CM-type distortion, showing that the proposed RTCS presents more reliable and high-spectral-spatial quality of the reconstructed HSI under stripe effect scenarios.

\begin{figure*}
    \begin{center}
    \includegraphics[width=1\textwidth] {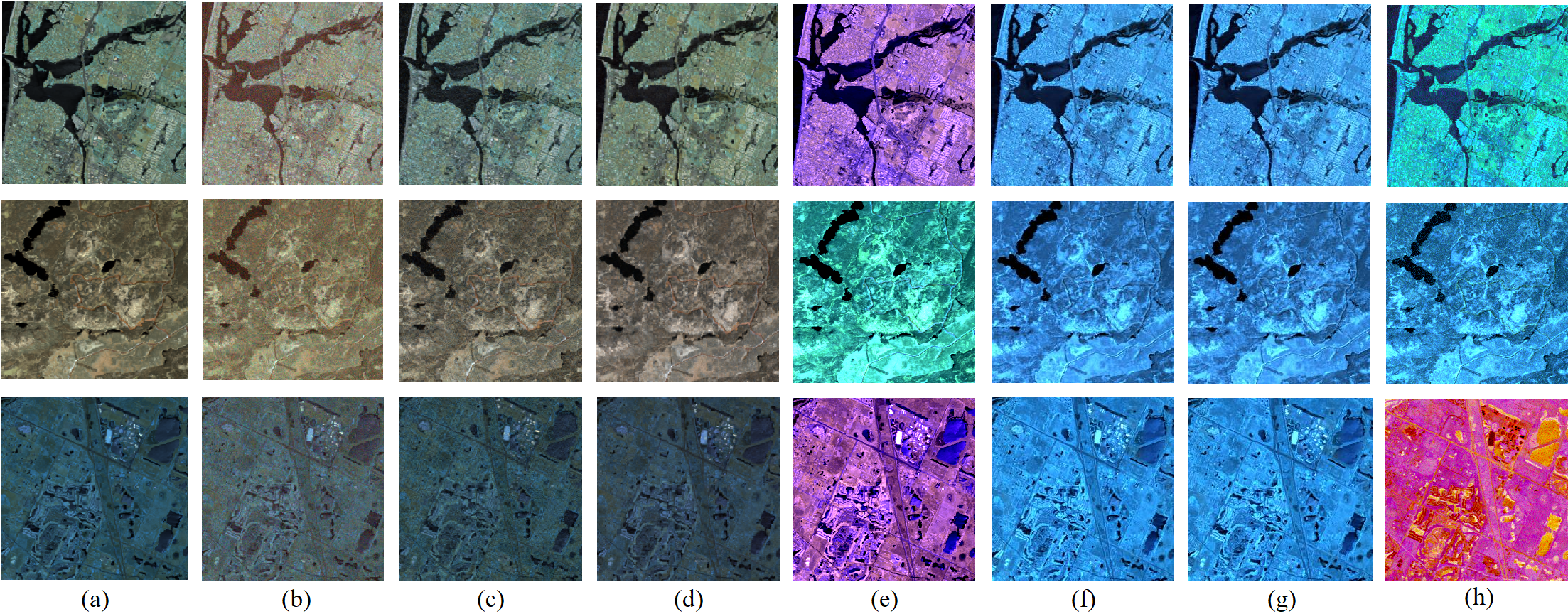}
    \end{center}
    \caption{The false color representation of the visualized hyperspectral image of the (a) ground truth, and the reconstructed counterpart under noised data with SNR = 15dB and sampling rate $\approx 1\%$, using (b) noised data (c) the proposed RTCS, (d) DCSN \cite{dcsn}, (e) TenTV \cite{wang2017compressive}, (f) AAHCS \cite{aahcsd}, (g) SPACE \cite{space}, (h) H-LSS \cite{martin2016hyperspectral}.}
    \label{fig:denoise}
\end{figure*}

As for visualized results comparison, the false RGB image of HSI could be synthesized by normalizing the band indices 13, 25, and 61. Figure \ref{fig:denoise} illustrates the false RGB image visualized from the reconstructed HSI of the different HCS methods under the 50-80 bands completely missing (i.e., CM), in which the red spectrum is also missing. As a result, the visualized results of the proposed RTCS are significantly better than that of other peer methods, implying the spectral signature of the reconstructed HSI using the proposed RTCS is promising. The state-of-the-art HCS methods, AAHCS \cite{aahcsd} and DCSN \cite{dcsn}, fail to reconstruct the visually pleasing results, implying the stripe effect could greatly influence the compressed signal $\bm Z$. Although the state-of-the-art inpainting method is adopted to recover the complete HSI from the corrupted one, the lower spatial and spectral quality cannot provide rich enough information for further remote-sensing-related analysis. Since the results reported in ADMM-ADAM \cite{admmadam} is evaluated in the pure and high-quality HSI with stripe effect, it is hard to guarantee such high-quality of the decoded HSI under a very low sampling rate (e.g., $\approx 1\%$), leading to unstable performance. In contrast, the proposed RTCS successfully reconstructs the missing parts from the compressed representation directly, leading to a promising visual quality of the decoded HSI and enabling the other downstream applications possible, even when some sensors are partly/completely broken.
As one can argue that the missing types could vary due to different types of sensors, the proposed RTCS requires to re-train the network, consisting of encoder and decoder, to enable the capability of HSI inpainting of the new missing type. Notably, the proposed RTCS is few-shot-compatible (cf. the following section), leading the cost of fine-tuning the RTCS for new types of stripe effects could be low and easy.

	\begin{table*}[ht]
		\vspace{0.2cm}
		\centering 
		\caption{{Performance comparison under different numbers of the training samples (\%) between the proposed RTCS and other HCS methods (sampling rate $\approx 1\%$).}}
		\scalebox{0.9125}{
			\begin{tabular}{|l|c|c c c c c |c|}\hline
			\multirow{2.2}{*}{Method} & \multirow{2.2}{*}{\shortstack{Training \\ samples}}

               & \multicolumn{6}{c|}{{{Test set}} (PSNR$\uparrow$ / RMSE$\downarrow$ / SAM$\downarrow$)} \\ \cline{3-8} 
			 && C-type	&	M-type &	F-type & L-type & O-type & Averaged \\ \hline 

			\multirow{2.2}{*}{DCSN\cite{dcsn}} & 10\%&	28.038/176.860/4.287 & 28.705/71.969/3.044 & 28.833/84.082/3.101 & 29.533/103.691/3.302 & 30.576/70.463/1.865 & 29.137/101.413/3.120 \\ \cline{2-8}
   			 & 20\%&	29.062/165.511/3.984 & 29.428/68.925/2.772 & 28.540/84.651/3.009 & 29.247/105.912/3.224 & 30.432/72.712/1.793 & 29.342/99.542/2.957 \\ \hline
            \multirow{2.2}{*}{RTCS} & 10\% &	\textbf{34.264/92.628/2.053} & \textbf{34.030/39.627/1.611} & \textbf{34.102/44.867/1.522 }& \textbf{34.763/51.488/1.522} & \textbf{35.664/42.866/0.932} & \textbf{34.565/54.295/1.528} \\ \cline{2-8}            
             & 20\% &	\textbf{34.457/89.947/1.999} & \textbf{34.347/38.111/1.516} & \textbf{34.528/42.481/1.430} &\textbf{ 35.322/48.756/1.385} & \textbf{35.990/41.314/0.887} & \textbf{34.929/52.122/1.444} \\ \hline 
            
		\end{tabular}}
		\vspace{0.2cm}
		\label{tab:fsl}
	\end{table*}

\begin{figure*}[t]
    \begin{center}
    \includegraphics[width=1\textwidth] {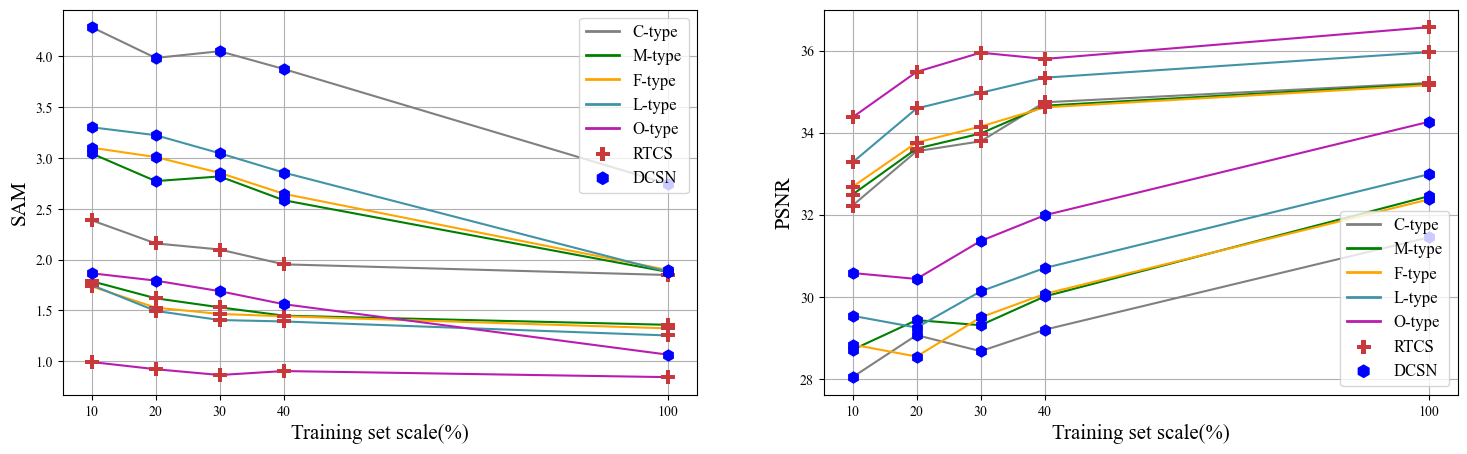}
    \end{center}
    \caption{The performance comparison of the DL-based HCS methods on varying amounts of the training samples.}
    \label{fig:fsl_curve}
\end{figure*}

\subsubsection{\textbf{Generalizability Evaluation}}
The data-driven approach usually suffers performance degradation issues from insufficient training samples. Compared to the optimization-based HCS methods, the data collection in the DL-based HCS methods is usually tedious and time-consuming. Since the number of parameters of the proposed RTCS is relatively small, the proposed RTCS should be able to train well using a relatively small-scale training set without overfitting. In this Section, we reduce the number of training data by randomly sampling $10\%$ to $40\%$ from the original training set (say, 1,000 HSI) to evaluate the performance of the same test set (say, 100 HSI). As shown in the Table \ref{tab:fsl}, the proposed RTCS could obtain the promising performance using only $10\%$ training samples, i.e., 100 HSI only, while the DCSN \cite{dcsn} requires $100\%$ training set to have similar performance. As claimed in the ADMM-ADAM theory \cite{admmadam}, a huge amount of the HSI collection is infeasible, so they have developed ADMM-ADAM using just small training data consisting of 400 HSI. However, we show that the proposed RTCS is more effective, even with only 100 HSI as the training set. As shown in the Figure \ref{fig:fsl_curve}, the proposed RTCS improves smoothly when the number of training samples increases to the $30\%$ of the original dataset, implying that the performance of our RTCS becomes promising using just a small amount of the training data. 

Besides data collection is time-consuming for DL-based HCS methods, however, the proposed RTCS only requires a tiny amount of the training samples, making our RTCS useful and practical. Once a new type of HSI sensor appears, the proposed RTCS requires only small data, making the HCS training and deployment more effective and efficient.

Our RTCS was developed on a proprietary dataset in the initial phase, laying the groundwork for a foundational model in hyperspectral imaging (HSI) compressed sensing. This foundational model facilitates rapid domain adaptation through fine-tuning with a limited subset of a target dataset—a method that proves to be computationally economical.

To evaluate the framework's performance, we engaged in further experimentation using an auxiliary dataset referenced as \cite{dcsn}. The RTCS framework was subjected to training on the designated training subset of this dataset, with its effectiveness subsequently gauged on the test subset. In an endeavor to validate the model's applicability, only 10\% of the \cite{dcsn} training samples were employed to fine-tune the pre-established RTCS model, which substantiated its efficacy on the test subset. The DCSN dataset \cite{dcsn} encompasses 2300 HSI samples, encapsulating a variety of landscapes, including City (C-type), Land (L-type), Farm (F-type), and Mountainous (M-type) terrains. Training the RTCS on the entirety of this dataset is projected to enhance performance; however, the collection of comprehensive datasets for novel target HSI applications is often constrained by practicality and resource availability.

The few-shot compatibility embedded within our RTCS enables the efficient training of the network on novel datasets with a minimal quantity of samples, mirroring the process of task-specific fine-tuning seen with foundational models. Illustrated in the Table \ref{tab:cmp_rtcs_data2}, the RTCS achieves state-of-the-art performance with a mere 1\% of the training samples, equivalent to 23 HSI images. This advancement highlights the significance and viability of Few-shot Compatibility in practical settings, where extensive data collection and model training are typically arduous. Crucially, this feature permits the swift and efficacious deployment of our RTCS across a spectrum of HSI types, employing only a handful of training samples and thereby promoting extensive adaptability and practicality for CubeSat HSI applications.

    \begin{table*}[t]
	\vspace{0.1cm}
	\centering 
	\caption{SNR performance comparison under sampling rate $\approx 1\%$ and $5\%$ between the proposed RTCS and other state-of-the-art HCS methods. The result in \textbf{bold} denotes the best method under the same sampling rate.}
	\scalebox{0.92}{
		\begin{tabular}{|l|c| c c c c |}\hline
			\multirow{2.2}{*}{Method} 
			  & \multirow{2.2}{*}{SR} & \multicolumn{4}{c|}{SNR (PSNR$\uparrow$ / RMSE$\downarrow$ / SAM$\downarrow$)} \\ \cline{3-6}
			 & & 25dB & 30dB & 35dB & 40dB \\ \hline
    
            \multirow{2.2}{*}{AAHCSD\cite{aahcsd}} & $1\%$ & 24.426 / 1181.978 / 7.574 & 27.231 / 1173.048 / 5.760 & 29.471 / 1169.741 / 4.908 & 31.016 /  1168.419 / 4.533 \\ \cline{2-6}
            & $5\%$ & 25.995 / 1455.559 / 9.553 & 28.296 / 1250.833 / 5.783 & 30.976 / 1177.976 / 3.880 & 33.420 / 1173.690 / 2.959 \\ \hline
            
            \multirow{2.2}{*}{SPACE\cite{space}} & $1\%$ & 24.880 / 151.329 / 6.730 & 27.539 / 88.070 / 3.149 & 28.616 / 112.770 / 4.843 & 29.456 / 108.174 / 4.585 \\ \cline{2-6}
            & $5\%$ & 28.452 / 91.104 / 4.404 & 32.060 / 58.909 / 2.847 & 34.875 /  44.118 / 2.087 & 36.576 / 38.130 / 1.755 \\ \hline

            \multirow{2.2}{*}{SpeCA\cite{martin2016hyperspectral}} & $1\%$ & 14.032 / 446.873 / 19.291 & 15.582 / 409.135 / 17.873 & 16.300 / 394.734 / 17.352 & 16.666 / 389.423 / 17.172 \\ \cline{2-6}
            & $5\%$ & 22.163 / 209.673 / 10.392 & 24.340 / 156.942 / 7.613 & 26.048 / 129.245 / 6.039 & 27.326 / 114.161 / 5.104 \\ \hline

            \multirow{2.2}{*}{TenTV\cite{wang2017compressive}} & $1\%$ &  18.486 / 258.820 / 10.333 & 18.498 / 237.224 / 9.359 & 18.321 / 229.991 / 9.024 & 18.248 / 226.777 / 8.875\\ \cline{2-6}
            & $5\%$ &  22.388 / 156.861 / 6.539 & 23.240 / 138.504 / 5.607 & 23.721 / 128.602 / 5.130 & 23.925 / 124.294 / 4.926\\ \hline

            \multirow{2.2}{*}{H-LSS\cite{zhang2016locally}} & $1\%$ & 22.564 / 264.779 / 8.367 & 22.726 / 246.871 / 8.227 & 22.900 / 268.792 / 8.335 & 22.861 / 250.298 / 8.185 \\ \cline{2-6}
            & $5\%$ & 26.245 / 137.323 / 5.631 & 27.318 / 123.489 / 5.244 & 27.803 / 118.771 / 5.042 & 28.129 / 118.016 / 4.976 \\ \hline

            \multirow{2.2}{*}{DCSN\cite{dcsn}} & $1\%$ & 31.308 / 80.501 / 2.382 & 32.153 / 74.007 / 2.079 & 32.501 / 71.657 / 1.959 & 32.635 / 70.858 / 1.915 \\ \cline{2-6}
            & $5\%$ & 36.301 / 50.942 / 1.029 & 36.820 / 48.958 / 0.899 & 37.009 / 48.325 / 0.847 & 37.075 / 48.119 / 0.828 \\ \hline

            \hhline{|=|=====|}
            \multirow{2.2}{*}{RTCS} & $1\%$ & \textbf{35.649 / 48.190 / 1.309} & \textbf{36.254 / 45.041 / 1.216} & \textbf{36.470 / 43.995 / 1.184} & \textbf{36.542 / 43.672 / 1.173} \\ \cline{2-6}
            & $5\%$ & \textbf{40.264 / 26.433 / 0.974} & \textbf{42.137 / 21.451 / 0.777} & \textbf{43.006 / 19.623 / 0.704} & \textbf{43.332 / 19.018 / 0.679} \\ \hline
	\end{tabular}}
	\vspace{0.1cm}
	\label{tab:snr_rtcsCR}
    \end{table*}

{\red
\begin{table*}[t]
	\vspace{0.1cm}
	\centering 
	\caption{Performance comparison between the proposed RTCS and other HCS methods under a sampling rate (SR) of $\approx 1\%$  for the dataset in \cite{dcsn}. The result in \textbf{bold} denotes the best method under the same sampling rate, and RTCS (1\%) denotes that the RTCS was fine-tuned by only 1\% training samples of the entire dataset in \cite{dcsn}.}
	\scalebox{0.80}{
		\begin{tabular}{|l|cc c c | c|}\hline
			\multirow{2.2}{*}{Method} 
			& \multicolumn{5}{c|}{Dataset in  \cite{dcsn} (PSNR$\uparrow$ / RMSE$\downarrow$ / SAM$\downarrow$)} \\ \cline{2-6} & C-type & M-type &	F-type & L-type  & Averaged \\ \hline
            AAHCS*\cite{aahcsd} & 30.259 / 169.090 / 6.767 & 31.496 / 37.446 / 3.617 & 31.340 / 59.915 / 4.375	& 28.768 / 49.172 / 4.733 & 31.121 / 46.418/ 3.950 \\ \hline
            
            SPACE*\cite{space} & 24.129 / 613.661 /7.207 & 29.161 / 140.415 / 3.743 & 29.674 / 64.151 / 3.121 & 27.727 / 209.757 / 4.446 & 28.855 / 158.751 / 3.892  \\ \hline

            SpeCA*\cite{martin2016hyperspectral} & 9.299 / 784.867 / 42.863 & 15.377 / 234.735 / 21.510 & 11.701 / 407.530 / 33.036 & 14.024 / 225.772 / 22.006 & 14.585 / 274.191 / 23.659 \\ \hline

            TenTV\cite{wang2017compressive} & 20.208 / 570.255 / 26.247 & 18.533 / 260.221 / 22.972 & 20.401 / 248.994 / 18.714 & 18.824 / 314.248 / 25.523 & 18.835 / 277.403 / 22.930 \\ \hline
  
            H-LSS\cite{zhang2016locally} & 20.798 / 228.967 / 9.241 & 23.942 / 93.679 / 6.206 & 21.552 / 73.796 / 9.898 & 23.578 / 331.702 / 8.761 & 23.518 / 123.890 / 7.016 \\ \hline
 
            DCSN\cite{dcsn}& 34.274 / 65.120 / 2.016 & 33.729 / 30.620 / 1.631 & 35.908 / 17.408 / 1.380 & 35.566 / 21.558 / 1.408 & 34.195 / 29.526 / 1.594 \\ \hline
            
            \hhline{|=|=====|}
             RTCS (1\% / 23 HSIs)  &34.278 / 66.858 / 1.849 & 32.499 / 33.231 / 1.830 & 35.694 / 13.562 / 1.418  & 33.984 / 35.138 / 1.530 & 34.114 / 37.197 / 1.657 \\ \hline
            RTCS (10\% / 230 HSIs)  & \textbf{35.160 / 62.948 / 1.656} & \textbf{34.074 / 29.049 / 1.471 }& \textbf{36.835 / 12.420 / 1.235} & \textbf{35.407 / 30.906 / 1.259} & \textbf{34.573 / 29.259 / 1.428} \\  \hline

	\end{tabular}}
	\vspace{0.1cm}
	\label{tab:cmp_rtcs_data2}
\end{table*}
}

    \begin{table*}[t]
	\vspace{0.15cm}
	\centering
	\caption{The run-time and computational complexity comparison of the decoders of the proposed RTCS and other methods under different computing platforms.}
	\scalebox{0.92}{
		\begin{tabular}{|l|cccc|c|c|c|} \hline
            \multirow{3}{*}{Method} & \multicolumn{7}{c|}{{{Computing platform}}} \\ \cline{2-8} &\multicolumn{4}{c|}{GPU (GTX 1660)} & CPU (i9-9900) & Jetson TX2 GPU & Denver 2 64-bit CPU \\ \cline{2-8}  & \multicolumn{1}{c|}{\textbf{\#Params}} & \multicolumn{1}{c|}{\textbf{FLOPs}}  & \multicolumn{1}{c|}{\textbf{Run-time}} & \textbf{PSNR$\uparrow$ / RMSE$\downarrow$ / SAM$\downarrow$} & \textbf{Run-time} & \textbf{Run-time} & \textbf{Run-time}\\ \hline
            AAHCS\cite{aahcsd} & \multicolumn{1}{c|}{-} & \multicolumn{1}{c|}{-} & \multicolumn{1}{c|}{-} & 32.978 / 107.917/4.287 & 274.15 & - & - \\ \hline
            SpeCA\cite{martin2016hyperspectral} & \multicolumn{1}{c|}{-} & \multicolumn{1}{c|}{-} & \multicolumn{1}{c|}{-} & 16.830 / 386.398 / 17.094 & 0.257 & - & - \\ \hline
            DCSN\cite{dcsn} & \multicolumn{1}{c|}{11.95M} & \multicolumn{1}{c|}{0.866G} & \multicolumn{1}{c|}{0.035} & 32.703 / 70.471 / 1.893 & 0.431 & 0.033 & 3.85\\ \hline
            DCSN-TVM\cite{dcsn} & \multicolumn{1}{c|}{11.95M} & \multicolumn{1}{c|}{0.866G} & \multicolumn{1}{c|}{0.023} & 32.703 / 70.471 / 1.893 & 0.383 & 0.025 & 2.119\\ \hline
            
            \hhline{|=|=|=|=|=|=|=|=|}
            
            RTCS & \multicolumn{1}{c|}{6.291M} & \multicolumn{1}{c|}{0.297G} & \multicolumn{1}{c|}{0.023} & 36.537 / 43.681 / 1.176 & 0.392 & 0.024 & 3.497 \\ \hline
            RTCS-TVM & \multicolumn{1}{c|}{6.291M} & \multicolumn{1}{c|}{0.297G} & \multicolumn{1}{c|}{0.011} & 36.537 / 43.681 / 1.176 & 0.243 & 0.019 & 2.011\\ \hline
            \end{tabular}}
	\label{tab:complexity}
    \end{table*}

\begin{table*}[t]
		\vspace{0.2cm}
		\centering 
		\caption{The performance comparison of the proposed RTCS with/without integer-8-based encoder.}
		\scalebox{0.82}{
			\begin{tabular}{|l| c c c c c |c|}\hline
			\multirow{2.2}{*}{Method} 
			& \multicolumn{6}{c|}{{{Test set}} (PSNR$\uparrow$ / RMSE$\downarrow$ / SAM$\downarrow$)} \\ \cline{2-7} & C-type & M-type &	F-type & L-type & O-type  & Averaged \\ \hline
                FP32 & 36.036 / 74.521 / 1.66 & 36.113 / 31.814 / 1.215 & 36.129 / 33.908 / 1.169 & 36.952 / 40.374 / 1.092 & 37.455 / 37.788 / 0.734 & 36.537/43.681/1.176 \\ \hline
                INT8 (QAT) & 34.992 / 84.096 / 1.973 & 34.914 / 36.454 / 1.434 & 34.801 / 40.819 / 1.472 & 35.353 / 48.415 / 1.400 & 34.891 / 46.213 / 1.024	 &  34.990/51.200/1.461 \\ \hline 
                INT8 (PAQ) &  34.952 / 84.748 / 1.993 & 34.844 / 36.734 / 1.448 & 34.714 / 41.221 / 1.492 & 35.280 / 49.002 / 1.419 & 34.858 / 46.616 / 1.038 	  &  34.929/51.664/1.478 \\ \hline

		\end{tabular}}
		\vspace{0.2cm}
		\label{tab:qat}
	\end{table*}

\subsubsection{\textbf{Transmission Noise}}

In order to evaluate the robustness of the proposed method, we follow the training process suggested in Denoised AutoEncoder\cite{dae1,dae2} and DCSN \cite{dcsn} to add the Gaussian noise with the 30dB Signal-to-Noise-Ratio (SNR) to learn the robust RTCS model. The rest hyperparameters of the proposed RTCS are identical to the default settings. Table \ref{tab:snr_rtcsCR} draws the performance comparison between the proposed RTCS and other state-of-the-art HCS methods under sampling rates $\approx 1\%$ and $\approx 5\%$. As a result, the proposed RTCS significantly outperforms the other HCS methods in terms of SAM, PSNR, and RMSE. Remarkably, the Robust DCSN (RDCSN) \cite{dcsn} also adopts the Gaussian noise during the training phase to make the model could robust to noise. However, the proposed RTCS still achieves a notable performance improvement, implying that the architecture of the proposed RTCS is better than that of DCSN \cite{dcsn} naturally. The results in the Figure \ref{fig:denoise} has shown the efficacy of the proposed RTCS. Compared with other HCS methods that have varying degrees of chromatic aberration, the reconstruction results of RTCS are closer to the ground-truth.

\subsection {Computational Complexity Analysis}\label{sec:complexity_analysis}
We compare the proposed RTCS to the other state-of-the-art HCS methods for computational complexity and run-time of encoding and decoding. Note that each result is measured with the same hardware. In the general computational comparison, a personal computer equipped with an i9-9900 CPU, 128GB Memory, and a low-end graphics processing unit (GPU) is used to measure the run-time of HCS for an HSI sized of $172\times 256\times 256$. Although the proposed RTCS is a stripe-like HCS, the run-time of the RTCS and DCSN \cite{dcsn} is measured in the whole HSI sized of $172\times 256\times 256$ since the optimization-based HCS methods work for the whole HSI only. In order to verify the efficiency of the proposed RTCS, we adopt NVIDIA Jeston TX2 to measure the run-time of the proposed RTCS and DCSN \cite{dcsn}. Note that the optimization-based HCS methods were developed based on Matlab, which is incompatible with ARM-based architecture; thus, those methods cannot run in Jetson TX2. We take the average of the run-time by repeatedly executing the inference 30 times. 

In the Table \ref{tab:complexity}, the proposed RTCS has the lowest parameter space as well as the computational complexity compared to other optimization- and DL-based approaches. In the run-time comparison, the proposed RTCS achieves the fastest inference time since the depth of the network is relatively shallow and the number of parameters is low, making real-time compressed sensing and decoding possible. Although AAHCS \cite{aahcsd} achieves good performance as illustrated in Section \ref{sec:objective_quality_assessment} under higher sampling rate $s_r \approx 5\%$, the decoding process of the proposed RTCS is 3000x faster than that of AAHCS \cite{aahcsd}, implying our RTCS is more feasible for real-time hyperspectral stripes decoding. Moreover, with the sampling rate $\approx 1\%$, the proposed RTCS significantly outperforms the peer methods for all quality indices, as well as achieves the lowest computational complexity. On the one hand, the encoder of the proposed RTCS only requires a matrix multiplication, making the cost of the computation and memory usage could be minimized in the miniaturized satellite. On the other hand, the decoder of the proposed RTCS could be deployed on edge devices due to its computational cheapness, which could greatly reduce the overhead of the central server and make more critical HSI applications possible.

More specifically, due to the real-time decoding requirement of the proposed RTCS, we deploy the proposed RTCS to the low-end device, NVIDIA Jetson TX2, with the TVM compiler \cite{chen2018tvm} for optimizing the efficiency of the DL-based approach (i.e., DCSN and our RTCS) for the specific hardware. Clearly, the proposed RTCS still achieves a very fast inference in the embedding platform, i.e., 19 Milliseconds (ms), while the DCSN in TX2 requires 25 ms for inference. In the DL-based HCS methods without TVM-optimization, the proposed RTCS still meets the real-time requirements (say, $1/0.024 = 41.6$ FPS), while DCSN is relatively hard to achieve such a fast inference process due to deeper network. Even in the ARM-based CPU in TX2, the inference time of the proposed RTCS is only 2 seconds, whereas the optimization-based HCS methods, including AAHCS \cite{aahcsd}, SpeCA \cite{martin2016hyperspectral}, and SPACE \cite{space}, failed to decode the HSI due to limited memory and insufficient computing power. Therefore, we could conduct that the proposed RTCS is efficient, as well as a state-of-the-art HCS method for miniaturized satellites.

We also further verify that the proposed RTCS is hardware friendly. The proposed encoder could deploy the HCS in the lightweight chip without any hardware accelerator, such as a GPU or intelligent processing unit (IPU), in a real-time sense. Although only one matrix multiplication is required in the encoder of the proposed RTCS, the floating-point with 32-bit precision (FP32) is still required, the same as the optimization-based HCS methods. Most low-end computing chips, such as Kneron KL720/FL520, require integer-based inference to maximize efficiency. Therefore, we further study the model quantization for our encoder to enable integer computation for saving power-consuming purposes by using quantization-aware training (QAT) and post-quantization (PQ) \cite{qat}. As shown in the Table \ref{tab:qat}, the quantized encoder of RTCS maintains promising performance while being more hardware-friendly."

Compared to the traditional optimization-based approach, the floating point operation is strictly required in the miniaturized satellite, leading to a higher computational cost inside the computational chip. As claimed in AAHCS \cite{aahcsd}, the basis could be calculated first and only all-addition operations required in the miniaturized satellite. However, the performance is unknown since the experiments were conducted on a data-dependant basis. In this fashion, the basis decomposition is still needed, so a number of floating point operations are required, which is computationally expensive. As a result, the encoder of the proposed RTCS could be state-of-the-art as well as computationally cheap.


     \begin{table}[t]
	\centering
	\caption{Ablation study under sampling rate $\approx 1\%$. }
 \scalebox{0.95}{
		\begin{tabular}{ |c| c| c| c | c | c | c |}
        \hline
		\textbf{L1} &\textbf{{{SAM}}} & \textbf{IMFA} & \textbf{FRDB}& \textbf{PSNR$\uparrow$} & \textbf{SAM$\downarrow$} &\textbf{Flops / \#Params} \\ \hline\hline
		 \checkmark& & \checkmark& &34.330 &  2.094 & 0.302G/6.434M \\ \hline
		 \checkmark& \checkmark& \checkmark& & 35.677 &  1.320 & 0.302G/6.434M  \\ \hline
		 \checkmark& \checkmark& & \checkmark& 34.837 &  1.435 & 0.291G/6.0753M \\ \hline
		 \checkmark& & & \checkmark& 35.037 &  1.936 & 0.291G/6.0753M \\ \hline
		 \checkmark& & \checkmark& \checkmark& 35.222 &  1.876 & 0.297G/6.291M  \\ \hline
         & \checkmark& \checkmark& \checkmark& 9.123 &  1.576 & 0.297G/6.291M  \\ \hline
		 \checkmark& \checkmark& \checkmark& \checkmark& \textbf{36.537} & \textbf{1.176} & 0.297G/6.291M \\ \hline
	\end{tabular}
 }
	\label{tab:ablation1}
    \end{table}	

\subsection{Ablation Study}
The proposed RTCS primarily consists of two distinct branches designed to aggregate features with varying equivalent receptive fields. In order to assess the efficacy of the dual-stream network architecture, we examine the performance of each individual stream within the proposed RTCS. We increase the number of base blocks, denoted as $n_f$, to 16 in order to bring the parameter space of each branch closer to that of the proposed RTCS. Table \ref{tab:ablation1} presents an ablation study of the proposed RTCS to validate the effectiveness of both the two-streamed architecture and the SAM loss.

In the first stream, the IMFA-Net-based HCS, which can be considered an enhanced version of DCSN \cite{dcsn}, demonstrates a high PSNR but limited spectral quality, as indicated by the SAM value. Comparing IMFA-Net with the original design of DCSN \cite{dcsn}, it becomes evident that more channels in the IMFA-Net decoder are necessary to restore the high-quality spectral signature from compressed HSI effectively. When the SAM loss is considered during the training phase, spectral quality is further improved, suggesting that the proposed SAM loss is better suited for preserving HSI signatures in HCS tasks. Owing to the simplified network of IMFA-Net, its performance surpasses that of DCSN, mainly when dealing with a small training set.

Likewise, the performance of the FRDB-Net-based HCS can be enhanced when both loss functions are utilized concurrently. In the proposed dual-stream architecture, significant performance improvements are achieved as a result of the judicious aggregation of multiple receptive fields. Although the number of convolutional kernels is considerably reduced through the use of grouped convolution, leading to diminished performance in the HSI spectral signature, the proposed SAM loss can guide network learning toward spectral preservation, resulting in a superior SAM index for the proposed RTCS.

In conclusion, the proposed dual-stream network architecture combined with the SAM loss demonstrates effectiveness, efficiency, robustness, and hardware compatibility, making it a valuable solution for the task at hand.

	\section{Conclusion}\label{sec:conclusion}
This study proposed an effective, efficient, and robust hyperspectral compressed sensing (HCS) network, termed real-time compressed sensing (RTCS), for significantly lower computational resources required in the encoder side at the miniaturized satellite, as well as hardware-friendly decoder inside the receiver based on edge devices.
  A very computationally cheap encoder with only one matrix multiplication with integer-8 precision, has been proposed to compressively sense the stripe-like hyperspectral image (HSI),  making HCS on CubSat possible and practical.
  A two-streamed-based decoder, consisting of two shallow and effective sub-networks, has been proposed to reduce the computational complexity to meet the requirements of edge devices by judiciously aggregating the middle-level and semantic features by the larger and regular convolutional kernel sizes for two sub-networks.

Benefiting from the lower number of parameters in the proposed RTCS, the proposed method only requires small data, making its high adaptation power for new datasets.
  A novel task-specific training policy with Spectral Angle Mapper-aware (SAM) loss has been proposed to improve the quality of the spectral signature preservation and enable the proposed RTCS to restore the complete HSI even if partial information of the HSI is missing or noisy. 
  Comprehensive experimental results demonstrated the superiority of the proposed RTCS. Moreover, the proposed RTCS is the first research to tackle the joint issue of the HSI restoration and compressed sensing in edge devices only, leading to our RTCS being practical and useful in real-world HCS applications.

	\bibliographystyle{IEEEtran}
	\bibliography{ref}	
\begin{IEEEbiography}[{\includegraphics[width=1in,height=1.25in,clip,keepaspectratio]{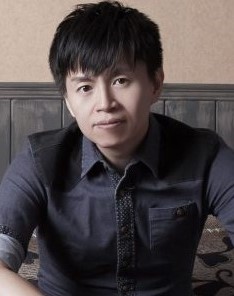}}]{Chih-Chung Hsu}
		
    (S'11-M'14-SM'20) received his B.S. degree in Information Management from Ling-Tung University of Science and Technology, Taiwan, in 2004, and his M.S. and Ph.D. degrees in Electrical Engineering from National Yunlin University of Science and Technology and National Tsing Hua University (NTHU), Taiwan, in 2007 and 2014, respectively.
    From 2014 to 2017, Dr. Hsu was a postdoctoral researcher with the Institute of Communications Engineering at NTHU. He served as an assistant professor with the Department of Management Information Systems at National Pingtung University of Science and Technology from February 2018 to 2021 and has been affiliated with the Institute of Data Science at National Cheng Kung University since 2021. His research interests primarily focus on computer vision and machine/deep learning, with applications in image and video processing. He has published several papers in top-tier journals and conference proceedings, including IEEE TPAMI, IEEE TIP, IEEE TMM, IEEE TGRS, ACM MM, IEEE ICIP, IEEE IGARSS, and IEEE ICASSP, including the best student paper award from the IEEE ICIP in 2019.
    Dr. Hsu became a Senior Member of the Institute of Electrical and Electronics Engineers (IEEE) in October 2020. He received the Best Young Professional Member Award from IEEE Tainan Section in 2023. Dr. Hsu leads the Advanced Computer Vision Lab (ACVLab), winning over 20 grand challenges at top-tier conferences over the years, such as the 3rd place award at the Learning to Drive Challenge from the IEEE International Conference on Computer Vision (ICCV), where he was also an invited speaker. He secured the 3rd place award in the Visual Inductive Priors for Data-Efficient Computer Vision Challenge and 1st place in the COV19D challenge at the European Conference on Computer Vision (ECCV) in 2020 and 2022. 
    
\end{IEEEbiography}

\begin{IEEEbiography}[{\includegraphics[width=1in,height=1.25in,clip,keepaspectratio]{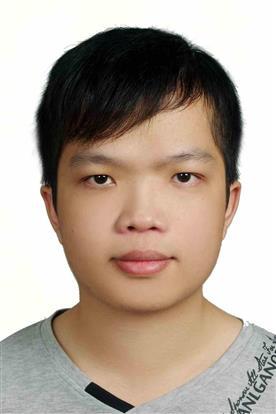}}]{Chih-Yu Jian}
    earned a bachelor's degree and graduated in 2018 from the Department of Information Management at Pingtung University of Science and Technology. He obtained his master's degree from the same department in 2020.
    From September 2020 to November 2022, he served as a master's researcher at Chang Gung Medical Foundation Hospital. Since December 2022, he has been working as a master's researcher at Far Eastern Memorial Hospital. His main interests are focused on computer vision and machine/deep learning, as well as the applications of medical imaging and natural language processing. He has published papers in IEEE ICASSP, IEEE VCIP, and NeuroImage Clinical.
    Over the years, he has won several significant challenges with his team at top conferences, such as securing third place in the Learning to Drive Challenge at the IEEE International Conference on Computer Vision (ICCV) and first place in the COV19D Challenge.		

\end{IEEEbiography}

\begin{IEEEbiography}[{\includegraphics[width=1in,height=1.25in,clip,keepaspectratio]{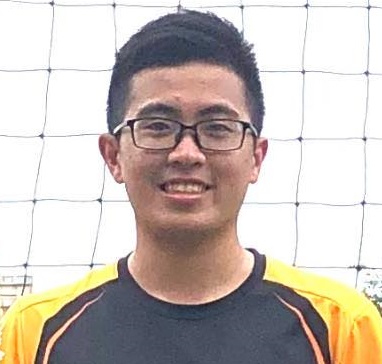}}]{Eng-Shen (James) Tu}
is a research assistant in the Software Engineering and Intelligent Test Automation Lab at National Cheng Kung University in Tainan, Taiwan. Currently, he is working towards his Bachelor's degree in Computer Science and Information Engineering at the same university. Though still early in his academic journey, James continues to explore and deepen his understanding of his chosen fields, always keen to contribute to new knowledge and innovative solutions in the domain of Data Science and Software Engineering.

\end{IEEEbiography}

\begin{IEEEbiography}[{\includegraphics[width=1in,height=1.25in,clip,keepaspectratio]{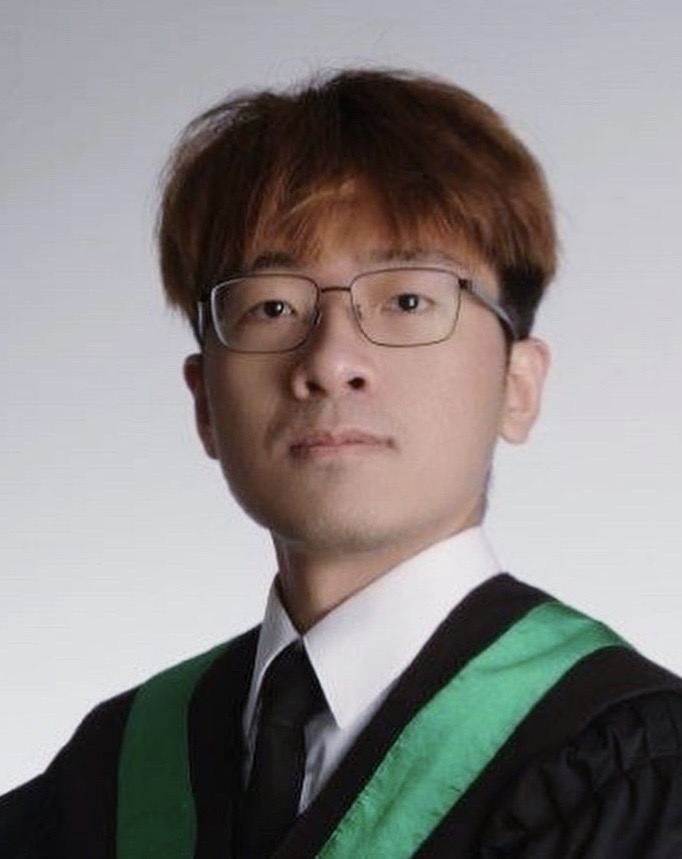}}]{Chia-Ming Lee}
received his B.S. degree at the Department of Statistics and Information Science, Fu Jen Catholic University (FJCU), Taiwan, in 2023. He is currently a M.S. student with the Advanced Computer Vision Laboratory (ACVLab) at the Institute of Data Science, National Cheng Kung University (NCKU), Tainan, Taiwan. His research interests mainly focus on computer vision, deep learning and their application.
  
    He have awarded the Jury Prize at Visual Inductive Priors Workshop from the IEEE International Conference on Computer Vision (ICCV). He secured the 1st place in the AI-enabled Medical Image Analysis Workshop and COVID-19 Diagnosis Competition held at the IEEE International Conference on Acoustics, Speech, Signal Processing (ICASSP). He received the Top Paper Award in the Social Media Prediction Challenge at ACM Multimedia (ACMMM).
\end{IEEEbiography}
	
\begin{IEEEbiography}[{\includegraphics[width=1in,height=1.25in,clip,keepaspectratio]{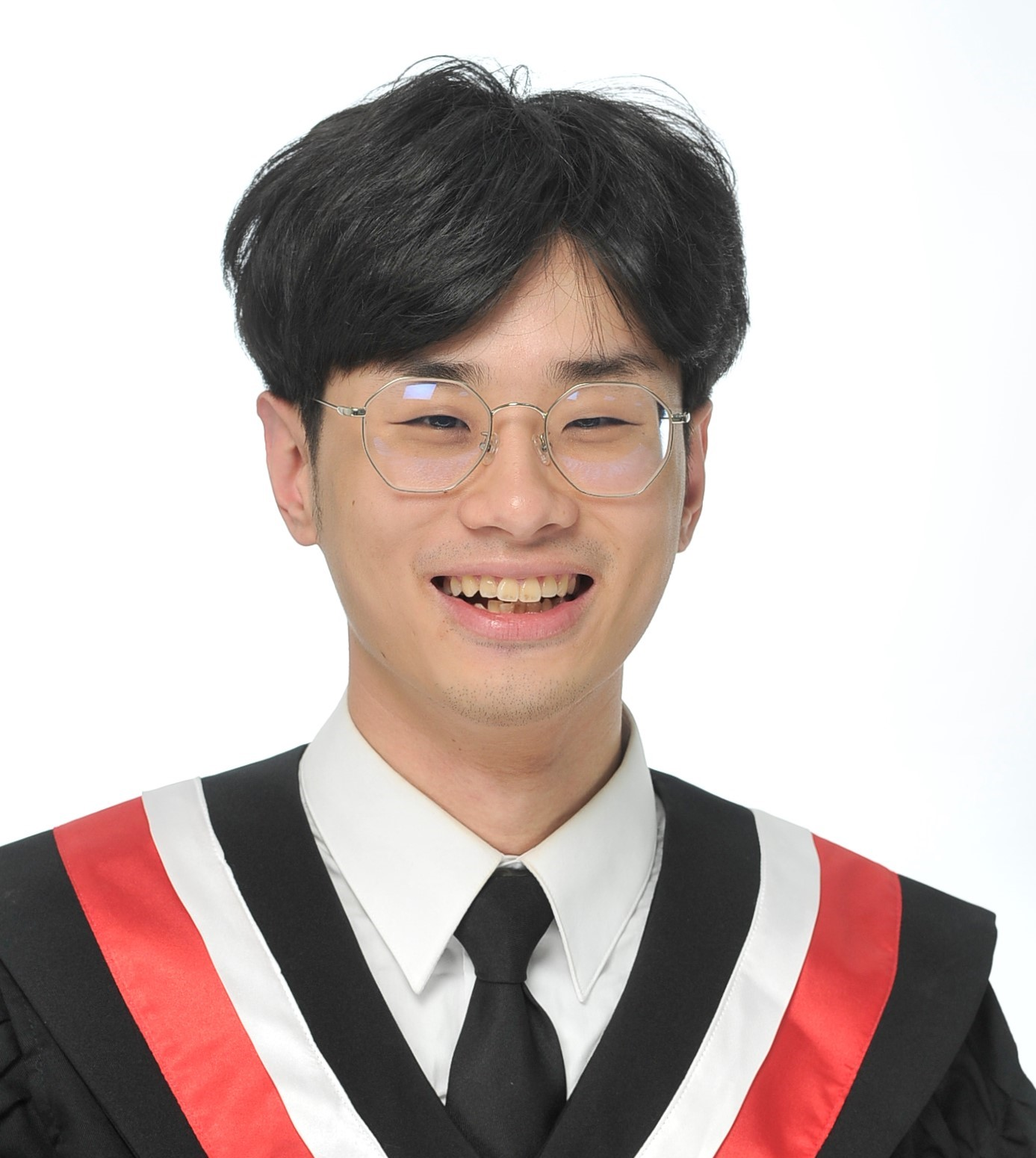}}]{Guan-Lin Chen}
received the B.S. degree from the Department of Financial Engineering and Actuarial Mathematics, Soochow University, Taipei, Taiwan, in 2020, and the M.S. degree from the Institute of Data Science, National Cheng Kung University, in 2022. He currently serves as an AI engineer at MobileDrive, Taipei, focusing on the application of learning-based methods for perception cameras. His research interests primarily focus on computer vision, machine learning, and deep learning, especially explainable and trustworthy AI.

\end{IEEEbiography}
	
\end{document}